\documentclass[]{article}

\usepackage{hyperref}
\usepackage{enumitem}
\usepackage{amsfonts} 
\usepackage{graphicx}

\usepackage{rotating}    

\usepackage[table]{xcolor}  
\usepackage{multirow}       
\usepackage{adjustbox}      
\usepackage{booktabs}       

\usepackage{subcaption}     
\usepackage{float}		    
\usepackage{pgfplots}
\usepgfplotslibrary{groupplots}

\usepackage{cite}          

\usepackage[normalem]{ulem}	

\usepackage{geometry}
\newgeometry{vmargin={25mm}, hmargin={34mm,34mm}}   

\usepackage[affil-it]{authblk}%

\usepackage{tikz}           
\usepackage{pgfplots}
\usetikzlibrary{patterns}
\usepackage{pgfplots,lipsum}

\usepackage{algorithm}
\usepackage{algpseudocode}
\usepackage{algcompatible}
\algrenewcommand\algorithmicindent{2.0em}%

\algdef{SE}{MyFor}{EndMyFor}[1]{%
	\textbf{foreach} #1}{%
	\textbf{end foreach}}

\usepackage{authblk}

\title{Text-Based Automatic Personality Prediction Using KGrAt-Net: A Knowledge Graph Attention Network Classifier}

\author[1]{Majid Ramezani}
\author[2]{Mohammad-Reza Feizi-Derakhshi}
\author[3]{Mohammad-Ali Balafar}
\affil[1,2]{Computerized Intelligence Systems Laboratory, Department of Computer Engineering, Faculty of Electrical and Computer Engineering, University of Tabriz, Tabriz, Iran}
\affil[3]{Department of Computer Engineering, Faculty of Electrical and Computer Engineering, University of Tabriz, Tabriz, Iran}
\affil[1]{Corresponding author: m\_ramezani@tabrizu.ac.ir}
\affil[2]{Corresponding author: mfeizi@tabrizu.ac.ir}
\affil[3]{balafarila@tabrizu.ac.ir}

\date{}

\begin{document}

\maketitle

\abstract{Nowadays, a tremendous amount of human communications occur on Internet-based communication infrastructures, like social networks, email, forums, organizational communication platforms, etc. Indeed, the automatic prediction or assessment of individuals' personalities through their written or exchanged text would be advantageous to ameliorate their relationships. To this end, this paper aims to propose \textit{KGrAt-Net}, which is a Knowledge Graph Attention Network text classifier. For the first time, it applies the knowledge graph attention network to perform Automatic Personality Prediction (APP), according to the Big Five personality traits. After performing some preprocessing activities, it first tries to acquire a knowing-full representation of the knowledge behind the concepts in the input text by building its equivalent knowledge graph. A knowledge graph collects interlinked descriptions of concepts, entities, and relationships in a machine-readable form. Practically, it provides a machine-readable cognitive understanding of concepts and semantic relationships among them. Then, applying the attention mechanism, it attempts to pay attention to the most relevant parts of the graph to predict the personality traits of the input text. We used 2,467 essays from the Essays Dataset. The results demonstrated that KGrAt-Net considerably improved personality prediction accuracies (up to 70.26\% on average). Furthermore, KGrAt-Net also uses knowledge graph embedding to enrich the classification, which makes it even more accurate (on average, 72.41\%) in APP.

\textbf{Keywords}: Automatic Personality Prediction (APP), Knowledge Graph (KG), Attention Network, Knowledge Graph Embedding, Knowledge Representation (KR), Big Five.
}

\section{Introduction}
\label{Sec:Introduction}
\textit{Personality} is the enduring set of traits and styles that an individual exhibits \cite{BergnerR2020}. It surrounds people's moods, attitudes, and opinions and it is explicitly expressed in their interactions with others \cite{Psychology2015}. Generally, it contains the comprehensive behavioural characteristics (both inherent and acquired) that can be observed in individuals' social relations and even in their relations with the environment. The word \textit{personality} stems from the Latin \textit{persona},  which indicates a mask that an actor wore in theatrical plays in the ancient world to represent and project the particular personality traits of the role.

Undoubtedly, Internet-based communications (such as various instant messaging and social networks, email, forums, organizational communication platforms, etc.) is increasing daily. Regarding their advantages and new challenges facing humankind (like COVID-19), various types of Internet-based communications are becoming increasingly ubiquitous. Being aware of individuals' personalities helps one perceive their thoughts, opinions, feelings, and responses to certain situations. It would be advantageous for everyone and may lead to better relationships, regardless of the type of relationship (the relationship between friends, the boss and employee, the seller and buyer, and others). Personality prediction or assessment has a broad spectrum of potential applications, for instance, different recommender systems, recruitment systems, online marketing, friend selection in social networks, personal counselling services, human resource management systems, etc.

\textit{Automatic Personality Prediction}, which hereafter we shall call \textit{APP}, is the assessment of human personality using computational approaches. It can be carried out based on the exchanged \textit{text} among people as a representation of human language, which is the primary communication tool among humankind. Several studies involving \textit{text-based} APP have exploited different methodologies for the purpose of APP in Internet-based communications. Preliminary studies of APP performed it using different linguistic features \cite{mairesse2007, Golbeck2011, Sumner2012, 2018Yuan, tighe2016perso} to capture more information about words and terms in the input texts, and finally use them as criteria to classify them. After a while, APP has received much attention and miscellaneous deep learning methods were utilized to acquire deeper information about text elements. Meanwhile, several attempts have been made to benefit the linguistic features and deep learning methods to improve the performance of APP \cite{Majumder2017, daSilva2018, YuanCu2018}. In recent years, to achieve more knowledgeable representations, investigators have examined the effects of embedding the text elements on APP \cite{2020Mehta, Ren2021, christian2021, JEREMY2021, ELDEMERDASH2020, Jiang2020, Wang2020Encoding}.

In general, perusing the studies, one can find that all of them are essentially trying to acquire more \textit{knowings} about the text elements, from linguistic feature-based methods to complex embedding-based deep learning approaches. That is to say, each of them by applying various methods, attempts to achieve a more knowledgeable representation of text elements to deal with rather than pure strings of characters. They are right since knowing is the basis of correct decisions and good performances for humans and machines. Besides, there is knowledge, and then there is knowing. Therefore, at first, we should discover the world behind the words. It will provide an excellent opportunity to promote the comprehension of text elements. This paper seeks to deal with this problem by acquiring a comprehensive, knowledgeable representation of the input text. Specifically, at first, it builds an equivalent graph for each text that entails all of the related knowledge of existing concepts in the input text. This graph (which we will refer to as \textit{knowledge graph}) is a machine-readable representation and a competent knowing-full substitution of the input text. Additionally, to achieve the most determining parts of the graph, after aggregating individual knowledge graphs into one aggregated graph, it is suggested to pay attention to the most critical parts of the graph (using \textit{attention mechanism}). Moreover, this paper suggests embedding the aggregated graph in another attempt to acquire an even more knowing-full representation of the graph elements. Finally, the resulting knowings will be utilized to build a classifier and perform APP.

Whilst, several researches have been carried out on text-based APP, no studies have been found that fundamentally are based on knowledge graph attention networks. This paper, for the first time (as far as we know), is focused on applying the attention mechanism over the knowledge graph in text-based APP. Indeed, a knowledge graph is a graph-based data model that formally represents the semantics of the existing concepts in the input text and models the knowledge behind them \cite{hogan2021}. Like all graphs, it is made up of nodes and edges in which the nodes represent the entities of the real world, and the edges connect pairs of nodes according to their relationship. It practically provides a representation of the information about the world in a form that a computer system can exploit to solve different tasks \cite{bergman2018}. Meanwhile, the graph attention mechanism is used to learn node representations which assign different importance to each neighbour's contribution \cite{velickovic2018graph}. In practice, an attention mechanism allows the model to focus on task-relevant parts of the graph and assists it in making better decisions \cite{Lee2019}. More specifically, applying the attention mechanism on graphs will yield some achievements, including it will adroitly allow the model to neglect and avoid the noisy parts of the graph \cite{LeeJ2018} which is a prevalent problem in knowledge graphs; it allows the model to assign a different relevance score to each node to pay attention to the most relevant information in current task \cite{velickovic2018graph} (a comprehensive description about graph attention networks, can be found in \cite{Lee2019}). To this end, this paper proposes \textit{KGrAt-Net}, which is a \textit{knowledge graph attention network text classifier}. It has a three-phase architecture including \textit{preprocessing}, \textit{knowledge representation}, and \textit{knowledge graph attention network classification}. After cleaning up the input text and transforming it to a more digestible form for the machine in the first phase, in the second phase, KGrAt-Net attempts to represent the entire knowledge in each of the input texts and build their corresponding knowledge graphs, using \textit{DBpedia} knowledge base. Finally, in the third phase, aggregating the individual knowledge graphs and applying an attention mechanism over the resulting graph, KGrAt-Net develops a model to classify the input text. To enrich the representation and improve classification, it also lets the users exploit knowledge graph embedding and graph attention network as an auxiliary representation.

This study aimed to address the following research questions:
\begin{itemize}
	\item[\bfseries RQ.1] How do the knowledge graph attention networks affect the performance of a text-based APP?
	\item[\bfseries RQ.2] How does enriching the knowledge graph attention network by knowledge graph embedding affect the performance of a text-based APP?
	\item[\bfseries RQ.3] Does classification using a knowledge graph attention network, affects equally the predictions in all personality traits? What about enriching it by knowledge graph embedding?
	\item[\bfseries RQ.4] How does the knowledge graph, as a representation of input text, affect the text-based APP?
	\item[\bfseries RQ.5] What obstacles stand in the way of APP when exploiting knowledge graph as the representation of input text?
\end{itemize}

This paper has been organized in the following way. The \hyperref[sec:PP]{second section} deals with personality assessment in psychology. A brief overview of the recent history of text-based APP is presented in the \hyperref[sec:literatureReview]{third section}. The \hyperref[sec:Methodology]{fourth section} is concerned with the methodology used for developing KGrAt-Net. At first, it describes the architecture of KGrAt-Net as a classifier, and then it explains how to use KGrAt-Net for APP. The \hyperref[sec:result]{fifth section} presents the findings of the research, and the \hyperref[sec:Discussion]{sixth section} includes a discussion of the implication of the findings as well as responses to the research questions. At last, some conclusions are drawn in the \hyperref[sec:conclusion]{seventh section}.

\section{Personality Assessment in Psychology}
\label{sec:PP}
Since the beginning of human interactions, people have likely been making personality assessments \cite{butcher2009}. Indeed, the personality of people has a substantial influence on their lives. Several psychological investigations have been carried out to demonstrate the influence of personality on various aspects of human life; for instance decision making \cite{MENDES201950}, innovativeness and satisfaction with life \cite{Ali2019}, good citizenship and civic duty \cite{PRUYSERS2019}, romantic relationship \cite{asselmann2020}, forgiveness \cite{walker2017exploring}, consumers' behaviours \cite{Liu2016ToBO}, job performance \cite{TISU2020job}, antisocial online behaviours \cite{MOOR2019}, suicidality \cite{moselli2021}, getting a pandemic like COVID-19 \cite{bacon2020}, and many others. They are fundamentally established to understand and appraise individual personality differences that characterize people for a variety of purposes during receiving psychotherapy services.

Psychologists usually attempt to appraise the personality of people by asking them to respond to a self-report inventory or questionnaire (please refer to \cite{vernon2014pers} for more detailed information). There are several personality models such as Big Five model \cite{soto2013bigfive}, Myers–Briggs Type Indicator (MBTI) \cite{Furnham2020mbti}, three trait personality model PEN \cite{RUCH2021PEN}, the sixteen personality factor (16PF) \cite{cattell2008sixteen}. Among them, the Big Five model is the most popular one which is widely used by both psychologists and computational researchers. It models human personality in five categories: \textit{Openness}, \textit{Conscientiousness}, \textit{Extroversion}, \textit{Agreeableness}, and \textit{Neuroticism}, known as the acronym \textit{OCEAN}, through assigning a binary value (true/false) to each of them. To clarify it, indicating some facets of each category may be helpful (please refer to \cite{ramezani2021, 2021ReviewFeizi} for more details):

\begin{itemize}
	\item \textbf{Openness (O)}: an inclination to embrace new ideas, arts, feelings, and behaviours; unconventional; focused on tackling new challenge; wide range of interests and so imaginative.
	\item \textbf{Conscientiousness (C)}: self-disciplined, well organized and dutiful; careful and hard-working; reliable, resourceful and on time.
	\item \textbf{Extroversion (E)}: outgoing, energetic, assertive and talkative; affectionate, sociable and articulate; enjoys being the centre of attention.
	\item \textbf{Agreeableness (A)}: an inclination to agree and accompany the others; altruist and unselfish; friendly, loyal and patient; modest, considerate and cheerful.
	\item \textbf{Neuroticism (N)}: an inclination to experience negative emotions like anxiety, anger, depression, sadness and envy; impulsive and moody; lack of confidence.
\end{itemize}

It also should be noted that the Big Five traits are mostly independent \cite{2019intToPsych}. That is to say, being aware of someone's one personality trait does not provide so much information on the remaining traits. All of the investigations in this study is relied on the Big Five personality model.

\section{Literature Review}
\label{sec:literatureReview}
During last years, by increasing the Internet-based infrastructures (like social networks), a large and growing body of literature has investigated the automatic personality prediction in miscellaneous contexts such as \textit{text}, \textit{speech}, \textit{image}, \textit{video}, and \textit{social media activities} (likes, visits, mentions, digital footprints, profile interpretation and etc.). Among these, \textit{text} as one of the most significant representations of human language, which is adroitly capable of reflecting the writer's personality \cite{2021MORENO}, has achieved considerable attention. 

Tracing the evolution of text-based APP systems, we can classify the previous studies into five categories: \textit{linguistic and statistical methods}, \textit{hybrid methods} (combination of linguistic methods and deep learning-based methods), \textit{embedding methods}, \textit{ensemble modeling methods}, and \textit{attention-based methods}

\textbf{Linguistic and statistical methods}: several attempts have been made to perform text-based APP. APP's first serious discussions and analyses emerged during the past two decades by exploiting linguistic and statistical knowledge of text elements. They primarily attempted to predict the personality of writers or speakers by assigning their words to pre-determined categories. Linguistic Inquiry and Word Count (LIWC) \cite{2001liwc} is one of the most widely used tools that counts the words in the input text and places them into several pre-defined linguistic and psychological categories. LIWC is a dictionary of words and word stems belonging to one or more pre-defined categories. Given a text, it simply computes the percentage of included words in each category. Since 2001, there have been different versions available until LIWC2015 \cite{LIWC2015}, which contains more than 89 categories. Mairesse features \cite{mairesse2007}, Medical Research Council (MRC) \cite{MRCdb}, Structured Programming for Linguistic Cue Extraction (SPLICE) \cite{SPLICE2012}, NRC Emotion Intensity Lexicon \cite{NRCAFFECTDic}, SenticNet \cite{2020SenticNet6}, etc. are other alternatives that provides linguistic features for words. Generally, the primary idea behind them is that the word usage in everyday language discloses individuals' thoughts, personalities and feelings.

Several researchers have reported applying linguistic and lexical features in APP \cite{2018Yuan, mairesse2007, Golbeck2011, Sumner2012, tighe2016perso}. One study by Yuan et al. \cite{2018Yuan} investigated the personality of the characters in vernacular novels. They assigned a vector for each dialogue using LIWC features, reflecting the characters' personalities. They mapped the vectors with Big Five personality traits and determined the personality labels. Similarly, Mairesse et al. \cite{mairesse2007} have utilized a variety of lexicon-based features in an effort to predict the personality traits from written text and spoken conversation. These features were also considered to predict the users' personality from Facebook text contents \cite{Golbeck2011} and Twitter posts \cite{Sumner2012}. Besides, some authors have also called into question the relationship between various mentioned features and different personality traits through determining their correlations \cite{Farandi2016, park2015, Schwartz2013}.

\textbf{Hybrid methods}: despite the previous superficial studies' approximate success in APP, there was a great tendency among researchers to focus on deep learning approaches to enhance APP systems' performance. Accordingly, a large and growing body of literature has investigated the combination of linguistic and statistical methods with deep learning methods \cite{2017linguisticDeep, Majumder2017, YuanCu2018}. Tandera et al. \cite{2017linguisticDeep} proposed a system to predict the personality of Facebook users in the Big Five model based on their information. They have combined several features like LIWC and SPLICE with a variety of traditional machine learning classification algorithms such as Naive Bayes, Support Vector Machine (SVM), Logistic Regression, Gradient Boosting, and Linear Discriminant Analysis (LDA), as well as some deep learning classification algorithms such as Long Short Term Memories (LSTMs), Gated Recurrent Unit (GRU), and 1-dimensional Convolutional Neural Network (CNN). They showed that applying deep learning methods can improve the performance of an APP system. In another study, Majumder et al. \cite{Majumder2017} proposed a CNN in which the document-level Mairesse features (extracted directly from each text) were fed into it. They have trained five separate identical binary classifiers for each five personality traits in the Big Five model. Yuan et al. \cite{YuanCu2018} investigated the combination of the LIWC features with deeper features that have been extracted through a deep learning model from Facebook status content. At first, they extracted the language features via the LIWC tool. Then, they added these features into a CNN that automatically extracts the deep features from textual contents to perform predictions. Practically, combining linguistic features with robust deep learning classifiers has ameliorated the prediction performance, but the efforts to investigate different deep learning-based methods were continued.

\textbf{Embedding methods}: a wide diversity of contributions has been published on deep learning-based APP, each of which has a distinct methodology. In recent years, an increasing number of studies have investigated the application of embedding methods to transfer the text elements from a textual space to a real-valued vector space \cite{2020Mehta, Ren2021, christian2021, ELDEMERDASH2020}. The authors in \cite{2020Mehta} integrated traditional psycholinguistic features such as Mairesse, SenticNet, NRC Emotion Intensity Lexicon, and VAD Lexicon (a lexicon of over 20,000 English words annotated with their valence, arousal and dominance scores), with several language model embeddings, including Bidirectional Encoder Representation from Transformers (BERT), ALBERT (A Lite Biomedical BERT) and RoBERTa (A Robustly Biomedical BERT Approach) to predict personality from the Essays Dataset in Big Five model. To that end, they have three classification algorithms: logistic regression, SVM and multilayer perceptron (MLP). In their investigations, Ren et al. \cite{Ren2021} proposed a multi-label APP model which combines emotional and semantic features (SenticNet), specifically sentiment analysis-based features, with sentence-level embeddings (produced by BERT). They believed that the current deep learning-based approaches suffer two limitations; the lack of sentiment information as well as psycholinguistic features and the lack of context information (context-based meaning of words). At last, they used three different models (CNN, GRU and LSTM) to perform final predictions on MBTI and the Big Five model. In an attempt to predict users' personalities in social media, Christian et al. \cite{christian2021} proposed a multi-modal deep learning system based on multiple pre-trained language models, including BERT, RoBERTa, and XLNet as the features extraction methods. Similarly, they also believed that the existing algorithms suffer from the absence of context information. They fed the embeddings into three independent feed-forward neural networks and finally performed the predictions by averaging the outputs. As a matter of fact, totally transferring the text elements from textual space to real-valued space in embedding methods yields better performance.

\textbf{Ensemble modeling methods}:‌ in the meantime, there have been a number of studies on how to take the advantages of several classifiers and benefit their prediction abilities simultaneously \cite{ramezani2021, ELDASH2021, Kunte2020, kazameini2020}. It was a great idea. The authors in \cite{ramezani2021} proposed an ensemble modelling method that was made up of five independent APP models, comprising: term frequency (TF) vector-based, ontology-based, enriched ontology-based, latent semantic analysis (LSA)-based, and a Bi-directional LSTM. Afterwards, they ensembled all the classifiers through a Hierarchical Attention Network (HAN) to predict the Big Five personality traits using the Essays Dataset. Perusing their investigation, one can find that they have exploited the specific abilities of inherently different classifiers and have put them together towards APP objectives. Specifically, they have exploited statistical and superficial information (TF vector-based method), hierarchical semantic information (ontology-based methods), co-occurrence-based and context-based information (LSA-based method), and sequence analysis-based information (BiLSTM-based method) concurrently to perform final prediction. In a similar attempt to benefit ensemble modelling advantages, El-Demerdash et al. \cite{ELDASH2021} proposed a deep learning-based APP system that was based on data-level and classifier-level fusion. They believed that fusion could improve the performance of an APP system. Therefore, exploiting the Essays Dataset and MyPersonality dataset together, as well as the pre-trained language models' embeddings like Elmo, ULMFiT, and BERT, they have performed the data-level fusion. Afterwards, for classifier-level fusion, they ensembled two multi-layer perceptrons (MLP) classifiers for BERT and Elmo along with Softmax classifier in the ULMFiT, to produce final predictions. Predictions were performed separately in each of the Big Five traits.

\textbf{Attention-based methods}: besides, a few more recent evidences \cite{Yang2021, Wang2021Attention, lynn2020} suggest the application attention mechanism in APP systems. Yang et al. \cite{Yang2021} investigated the multi-document personality prediction. The main idea behind their study was that most researchers in APP perform the predictions merely based on a single input document (or post) from an individual, whilst unifying them may conclude better predictions since some information appears when the documents are considered together rather than separately. To this end, they proposed a multi-document transformer that allows the encoder to access an individual's information in other documents (posts). In their seminal study, they pinpointed that different personality traits truly demand different contexts to let them be expressed. Therefore, trying to predict different traits of an individual from a single document may inherently cause some miss-predictions. Finally, they used an attention mechanism on top of the transformer to perform predictions using Kaggel and Pandora MBTI datasets. Wang et al. \cite{Wang2021Attention} proposed HMAttn-ECBiL, a hierarchical hybrid model based on a self-attention mechanism to acquire deep semantic information from input texts in MyPersonality dataset. HMAttn-ECBiL consists of two separate attention mechanisms: a CNN-based and a Bi-directional LSTM-based mechanism, respectively responsible for extracting the local features of text elements and sentence-level features. Their outputs, and the embeddings of the individual words were concatenated and then fed into a fully connected layer to perform predictions. They have attempted to improve the prediction performance through relying on diversity of features. In another study, Lynn et al. \cite{lynn2020} proposed a hierarchical attention mechanism to find the relative weight of users’ Facebook posts and then use them in predicting their personality in the Big Five model. To this end, they first used a GRU to acquire the message-level encodings. Then, a word-level attention mechanism was used to learn the weights of each word in the messages. Next, the message-level encodings and word-level attention were fed into a second GRU to yield user-level encodings. After that, a message-level attention mechanism acquires the weights of each user’s posts. Finally, the user encodings and the message-level attentions, were fed into two fully connected layers to perform the final prediction. Truthfully, considering the attention-based contributions provides evidence that could justify their competency in APP systems. 

Detailed analysis of contributions to text-based APP systems reveals that most of them try to acquire a meaningful and knowledgeable representation of input text elements to perform their predictions. Ranging from superficial linguistic methods to miscellaneous deep learning-based methods. In the meantime, there is a competent option for representing the complete knowledge of and knowledge among existing concepts in the input text, namely the knowledge graph. Although miscellaneous researches have been carried out on APP, only one study has attempted to investigate the application of knowledge graphs in APP systems. The authors in \cite{ramezani2022KGEnabeled} proposed a knowledge graph-enabled text-based APP system, relying on the Big Five model. Specifically, they first tried to build the corresponding knowledge graph for a given text using DBpedia knowledge base. Then, they enriched the graph using some linguistic and semantic lexicon, such as MRC, NRC and DBpedia ontology. Afterwards, they embedded the knowledge graph to acquire an even more meaningful and knowledgeable representation. Finally, the graph embeddings were fed into several independent deep learning models, namely CNN, RNN, LSTM and BiLSTM to perform the predictions using the Essays Dataset. Their study provides additional support for the importance of competent representation of the input text in APP systems.

Overall, the studies presented thus far provide evidence that outlines the critical role of input text representation as well as an adroit method to utilize it towards achieving the objectives of an APP system. Despite all of the studies, no one as far as we know has investigated APP using knowledge graph attention networks. That is to say, what is not yet clear is what would the effect of applying an attention mechanism on a comprehensive and knowledgeable representation of input text, namely the knowledge graph.

\section{Methods}
\label{sec:Methodology}
In this section, to comprehend the proposed APP method, at first the architecture of KGrAt-Net as a knowledge graph attention network classifier is meticulously described. Then, automatic personality prediction using KGrAt-Net is explained thoroughly.

\subsection{KGrAt-Net Architecture}
\label{sec:KGRATNET}
\begin{figure}
	\centering
	\includegraphics[width=\textwidth]{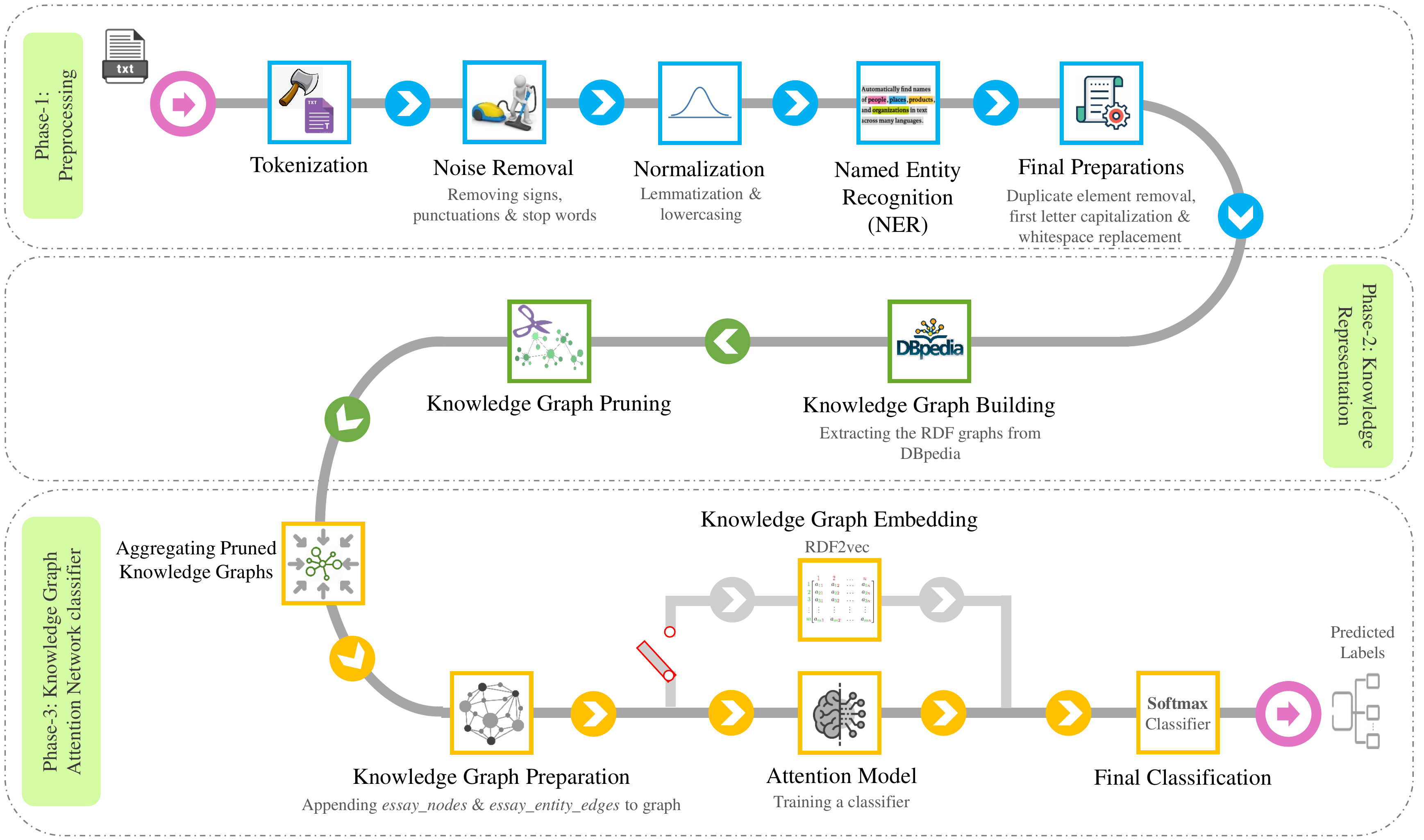} 
	\caption{The general architecture of KGrAt-Net}
	\label{Fig: General architecture}
\end{figure}

For the purpose of answering the research questions, which are stated in \hyperref[Sec:Introduction]{Introduction}, we suggested KGrAt-Net, a three-phase text classification approach which is basically founded on a knowledge graph attention network. \hyperref[Fig: General architecture]{Figure~\ref*{Fig: General architecture}} depicts the general architecture of KGrAt-Net.

\subsubsection{Phase-1: preprocessing}
This traditionally popular and prominent phase in natural language processing strives to clean up the input text and transform it into a more digestible form for the machine. It practically facilitates and enhances the main processes in the subsequent phases. Depending on the task, it may consist of miscellaneous activities. What follows is a complete description of preprocessing activities which were carried out chronologically by KGrAt-Net, as presented in \hyperref[Fig: General architecture]{Figure~\ref*{Fig: General architecture}}.

\paragraph{Tokenization}~\\
\textit{Tokenization} is the process of chopping up a text into pieces called \textit{tokens} which roughly correspond to words \cite{schutze2008}. These pieces are considered the smallest semantic units in text processing. To this end, we used the tokenizer which is provided by \textit{Natural Language Toolkit} (\textit{NLTK}) \cite{nltk}. The input of this step is the raw text, and the output is the list of its tokens.

\paragraph{Noise Removal}~\\
For the purpose of acquiring more plain data, it is highly preferred to remove undesirable and interfering pieces of input text. In current study \textit{signs}, \textit{punctuations} and \textit{stop words} were considered as undesirable pieces of text. To remove these pieces in the second step of preprocessing phase, we used NLTK.

\paragraph{Normalization}~\\
\textit{Normalization} is the process of canonicalizing tokens to a more uniform sequence so that matches occur, despite superficial differences in the character sequences of the tokens \cite{schutze2008}. In practice, it decreases the amount of information the machine has to deal with. Namely, the information that conceptually is similar but morphologically is different. In the third step of preprocessing phase, we performed \textit{lowercasing} and \textit{lemmatization} respectively, to normalize the input data. 

\textit{Lemmatization} is the morphological analysis of words that groups together their inflected forms and returns their bases or dictionary forms of them, which are called \textit{lemma}. An alternative option to reduce the inflected forms of words is \textit{stemming} which yields the words' \textit{stems}. It usually is fulfilled through chopping off the ending characters of the words, which often concludes the incorrect and misspelt forms of words. Therefore, it can be stated that the results of lemmatization (namely lemmas) are meaningful concepts rather than stems, in which we can find an equivalent match in the real world for them; what is really important in knowledge representation (as we will see in the second phase). This is why we preferred lemmatization to stemming. In the present study, the lemmatization was also carried out using NLTK.

\paragraph{Named Entity Recognition (NER)}~\\
\textit{Named entities} refer to the ``words" or ``sequences of words", which are the names of things such as organization, person, location, company, product, event and etc. Knowledge representation (and its result, namely knowledge graph), is actually the basis of KGrAt-Net's decisions. During knowledge representation, one should query all of the existing entities in the input text in a knowledge base. Indeed, knowledge bases basically entail the knowledge about the \textit{entities} of the real world. Hence, in an attempt to acquire the knowledge behind the words (or sequence of words) and represent them, it is absolutely critical to recognize the existing named entities in the input text. In this study, the \textit{spaCy} NER \cite{spacy} was used to recognize named entities. 

NER is the fourth step in preprocessing phase that yields a ``list" of entities and concepts which convey the fundamental notions that have appeared in the input text.

\paragraph{Final Preparations}~\\
Some further activities were performed in fifth step as the final preparations in preprocessing phase, including \textit{duplicate element removal}, \textit{first letter capitalization} and \textit{whitespace replacement}. Duplicate elements (entities and concepts) will impose further unnecessary computations on the system during the main process. To avoid it, we removed duplicate elements from the resulted list in the fourth step and provided a ``set" of entities and concepts. Furthermore, first letter capitalization, as well as whitespace replacement with an underscore for multi-word named entities, were done to prepare them to be matched with DBpedia knowledge base entries in the next phase.

A quick overview of the preprocessing phase, as can be seen in \hyperref[Fig: General architecture]{Figure~\ref*{Fig: General architecture}} can be concluded as follows:

\begin{enumerate}[label=(\roman*)]
	\setlength{\itemsep}{0pt}
	\setlength{\parskip}{0pt}
	\item input: the raw input text;
	\item output: a set of extracted entities and concepts from the input text;
	\item objective: to prepare a more digestible form of input text for main processes in next phases.
\end{enumerate}

\subsubsection{Phase-2: knowledge representation}
As explained in \hyperref[Sec:Introduction]{Introduction}, knowledge representation is the foundation of KGrAt-Net's decisions. In practice, the result of such representation, namely the knowledge graph, establishes the needed infrastructure for main processes. KGrAt-Net in this phase, at the first step, tries to represent the input text throughout building the knowledge graph, and then in the next step, to reduce the computational complexity and remove unessential parts, attempts to prune the knowledge graph.

\paragraph{Knowledge Graph Building}~\\
\label{sec:KGBuilding}
Indisputably, the final output of the first phase, namely the set of entities and concepts, fundamentally arranges the presented notions in the input text. In this step, KGrAt-Net attempts to acquire the general knowledge organization, which is provided by the input text. There is always knowledge behind every entity and concept that carries valuable information. Besides, some semantic relations among entities and concepts evolve and expand the knowledge organization. For this purpose, KGrAt-Net relies on \textit{DBpedia} knowledge base. In practice, it queries all of the elements of resulted set from the previous phase in DBpedia and extracts all the relevant knowledge of each entity and concept. In simple words, by querying each entity and concept, also known as \textit{resource}, KGrAt-Net asks for a complete description of them from DBpedia.

In consequence of such queries, KGrAt-Net acquires any resources related to the queried one (for more information about how to query in such knowledge bases, please refer to \cite{hogan2020sparql}). More specifically, a set of \textit{RDFs} is returned in response to each query. RDF stands for \textit{Resource Description Framework}, which is basically a standard for representing information on the Web. An RDF is a \textit{triple} which is made up of (\textit{subject, predicate, object}). Both of the \textit{subject} and \textit{object}, denote the resources and the \textit{predicate} denotes traits or aspects of the resource. In fact, a triple would be considered as a statement which declares the relationship (through the predicate) between two resources (subject and object). For instance, (\textit{the Mona Lisa, was created by Leonardo da Vinci}) is a sample RDF. Visually, a triple is suitably equivalent to a \textit{directed edge} in a graph from the subject toward the object, having the predicate as its label. More detailed information about RDFs is provided in \cite{RDF}.

The resulting set of RDFs, also known as \textit{RDF graph}, represents the knowledge of entities and concepts in a directed heterogeneous labelled multi-graph. RDF graph is wildly known as \textit{knowledge graph} (\textit{KG}). Therefore, at this step, we have a knowledgeable representation of the input text, namely the knowledge graph.

It has to be pointed out that KGrAt-Net applies two changes over the resulted knowledge graph to facilitate the main process in the next step, in particular: $i)$ it removes all of the labels (predicates) from the edges, $ii)$ it converts the multi-graph to a simple graph (through unifying multiple edges).

\paragraph{Knowledge Graph Pruning}~\\
\label{sec:KGPruning}
The extensive coverage of knowledge by the knowledge graphs brings new challenges for knowledge graph-based systems \cite{KGPruning2018, 2019KGpruning}. Very high computational complexity (for both time and memory) always is a significant obstacle that makes working with knowledge graphs extremely cumbersome or even sometimes impossible (regarding hardware constraints). Moreover, the extensive coverage of knowledge of concepts will bring some unessential knowledge that in practice would deteriorate the performance of the system in the current task \cite{KGPruning2018, 2019KGpruning}. Therefore, pruning the knowledge graph would be so helpful.

To perform knowledge graph pruning, KGrAt-Net uses a simple strategy; it just keeps those edges (RDFs) in which their corresponding subjects and objects (namely the source and destination nodes) have appeared in the input text. Considering that the resulting graph plays a leading role in KGrAt-Net and generally all of the knowledge graph-based systems, pruning the knowledge graph would be an open question.

At last, a quick overview of the second phase, as it can be seen in \hyperref[Fig: General architecture]{Figure~\ref*{Fig: General architecture}} can be concluded as follows:

\begin{enumerate}[label=(\roman*)]
	\setlength{\itemsep}{0pt}
	\setlength{\parskip}{0pt}
	\item input: the set of extracted entities and concepts from the input text;
	\item output: pruned equivalent knowledge graph of the input text;
	\item objective: acquiring a knowing-full representation of the input text, through building and pruning its equivalent knowledge graph.
\end{enumerate}

\subsubsection{Phase-3: knowledge graph attention network classification}
The needed infrastructure to implement a knowledge graph-based attention network is available at this point. In this phase, KGrAt-Net at first aggregates all of the pruned knowledge graphs and acquires an aggregated knowledge graph. Then it focuses on preparing the aggregated knowledge graph for text classification and finally designs a classification model by applying an attention network over the aggregated knowledge graph. It also equipped the model to utilize the aggregated knowledge graph embedding to enrich the model and acquire more accurate predictions.

\paragraph{Pruned Knowledge Graphs Aggregating}
\label{sec:Knowledge Graph Aggregation}~\\
To perform text classification over the pruned knowledge graphs, we need to aggregate them and build an \textit{``aggregated knowledge graph"}. In this step, all of the outputted pruned knowledge graphs from the second phase were aggregated, and a single aggregated knowledge graph was built. To avoid misunderstanding the concept of a knowledge graph, the term ``aggregated knowledge graph" hereafter is referred to as ``knowledge graph".

\paragraph{Knowledge Graph Preparation}
\label{sec:Knowledge Graph Preparation}~\\
At this step, KGrAt-Net tries to make the final preparations for text classification over the knowledge graph. Let's find out what kind of preparations are needed by KGrAt-Net at this point.

Perusing the research questions, KGrAt-Net developed a text classification model based on attention networks over the knowledge graph. Specifically, it attempts to develop a deep learning model over a graph. Hence, it is dealing with a \textit{Graph neural networks} (\textit{GNNs}) problem. GNNs are a general framework for defining deep neural networks on graph data \cite{2020GraphRepLear}. In general, GNNs provide three types of prediction tasks on graphs \cite{Book2022GNN}, including:

\begin{enumerate}[label=(\alph*)]
	\setlength{\itemsep}{0pt}
	\setlength{\parskip}{0pt}
	\item \textit{node-level} task: refer to those tasks associated with predicting some property of individual nodes in graph; such as node classification and node regression.
	\item \textit{edge-level} task: refer to those tasks associated with predicting the property of a pair of nodes in the graph; a common example of an such task is link prediction.
	\item \textit{graph-level} task: refer to those tasks associated with predicting some property of the whole graph; such as graph classification and graph property prediction. 
\end{enumerate}

A node classification task intends to train a model to predict the labels of the nodes \cite{2020nodeClassification}. KGrAt-Net figures out the text classification as a node classification problem. For this purpose, KGrAt-Net makes some changes to the knowledge graph. The current knowledge graph consists of a set of nodes as well as a set of edges; in which the nodes' set is the collection of extracted entities from DBpedia, which will be called hereafter \textit{entity\_nodes} and the edges' set is the collection of the semantic relationships among the extracted entities, that will be called hereafter \textit{entity\_entity\_edges}. At this step, KGrAt-Net appends \textit{essay\_nodes} to the nodes' set, each of which denotes an essay in the dataset. In fact, KGrAt-Net deals with \textit{essay\_nodes} to perform classification. Now we need to connect the \textit{essay\_nodes} to those related \textit{entity\_nodes} in knowledge graph. Hence, KGrAt-Net appends \textit{essay\_entity\_edges} to the edge's set, that relates the \textit{essay\_nodes} to \textit{entity\_nodes}. To do so, while appending an \textit{essay\_node}, KGrAt-Net checks out the occurrence of each \textit{entity\_node} in graph in current essay and then in case of finding an occurrence, it draws an \textit{essay\_entity\_edge} between the \textit{essay\_node} and the \textit{entity\_node}. Now, KGrAt-Net is ready to perform node classification.

In simple terms, KGrAt-Net performs text-based automatic personality prediction as a node classification problem over the knowledge graph (acquired through aggregating all individual pruned knowledge graphs). After appending the \textit{essay\_nodes} to this knowledge graph (each of which corresponds to one essay in Essays Dataset) as well as the edges among them and \textit{entity\_nodes}, the knowledge graph is prepared for classification. Hence, it should be declared that KGrAt-Net classifies the \textit{essay\_nodes} to perform text (node) classification. Namely, it predicts the binary label of each \textit{essay\_node} for each of the five personality traits. Considering the essence of the attention mechanism over the graph, KGrAt-Net utilizes the overall structure of the knowledge graph to perform the classification.

\paragraph{Knowledge Graph Attention Model}
\label{sec:Attention Network}~\\
In the present step, KGrAt-Net efforts to develop an attention network to perform node classification. Our experimental set-up to perform the attention network, bears a close resemblance to that proposed by Velickovic et al. \cite{velickovic2018graph}, but here the graph structure is different (since it is actually a knowledge graph and it encompasses \textit{essay\_nodes} as well as \textit{entity\_nodes}) besides that, it is used for text classification.
	
KGrAt-Net uses several \textit{graph attention layers}, each of which with separate learnable weights. A set of node features $h = \{\overrightarrow{h}_1, \overrightarrow{h}_2, ..., \overrightarrow{h}_N\}, \overrightarrow{h}_i \in \mathbb{R}^F,$ is the input of each single layer; where $N$ stands for the number of nodes, and $F$ stands for the number of features in each node. The output of the layer is a new set of node features, namely $h^\prime = \{\overrightarrow{h}^\prime_1, \overrightarrow{h}^\prime_2, ..., \overrightarrow{h}^\prime_N\}, \overrightarrow{h}^\prime_i \in \mathbb{R}^{F^\prime}$. The following descriptions, explain what happen in each single layer.

Being more specific, at first via a feature transformation process the input features ($\overrightarrow{h}_i$) are encoded to a higher level and dense features ($\overrightarrow{h}^\prime_i$). For this purpose, simply a \textit{weight matrix} of learnable parameters adopted for feature transformation, namely $\textbf{W} \in \mathbb{R}^{F^\prime \times F}$, is applied to each node ($\textbf{W}\overrightarrow{h}_i$).

Next, the \textit{attention scores} ($e_{ij}$) are calculated between all two neighbours using \textit{self-attention} which is a shared attentional mechanism $a : \mathbb{R}^{F^\prime} \times \mathbb{R}^{F^\prime} \to \mathbb{R}$, as revealed in \hyperref[eq:self-attention]{Equation~\ref*{eq:self-attention}}. That is to say, the attention score (namely $e_{ij}$) is calculated for those $j$ nodes, in which $j \in \mathcal{N}_i$, where $\mathcal{N}_i$ is the one-hop neighbours of the node $i$. In practice, it specifies the importance of the neighbours nodes (\textit{j}-th node) for current node (\textit{i}-th node). 

\begin{equation}
	\label{eq:self-attention}
	e_{ij} = a(\textbf{W}\overrightarrow{h}_i , \textbf{W}\overrightarrow{h}_j)
\end{equation}

Then for the purpose of normalizing the attention scores, softmax function is applied on each node's incoming edges, as shown in \hyperref[eq:softmax-normalization]{Equation~\ref*{eq:softmax-normalization}}. It causes the conversion of the output of the previous step into a probability distribution. Thus, the attention scores would be more comparable among other nodes. 

\begin{equation}
	\label{eq:softmax-normalization}
	\alpha_{ij} = softmax_j(e_{ij}) = {exp(e_{ij}) \over \sum_{k \in \mathcal{N}_i} {exp(e_{ik})}}
\end{equation}

Here, the attention mechanism (namely \textit{a}) is a single-layer feedforward network which is parametrized by a weight vector $\overrightarrow{\textbf{a}} \in \mathbb{R}^{2{F^\prime}}$, and followed by LeakyReLU activation function (with negative input slope $\alpha = 0.2$) to apply non-linearity in the transformation. The \hyperref[eq:attention-scores]{Equation~\ref*{eq:attention-scores}}, expands out the computations of attention scores.

\begin{equation}
	\label{eq:attention-scores}
	\alpha_{ij} = {exp\bigg( LeakyReLU \Big( \overrightarrow{\textbf{a}}^T \big[ \textbf{W}\overrightarrow{h}_i \parallel \textbf{W}\overrightarrow{h}_j \big] \Big) \bigg) \over \sum_{k \in \mathcal{N}_i} {exp\bigg( LeakyReLU \Big( \overrightarrow{\textbf{a}}^T \big[ \textbf{W}\overrightarrow{h}_i \parallel \textbf{W}\overrightarrow{h}_k \big] \Big) \bigg)}}
\end{equation}
where, the $\parallel$ denotes to concatenation process, and $^T$ denotes the transpose of the vector.

Having the attention scores, the corresponding linear combination of features are calculated to provide the final output features ($\overrightarrow{h}^\prime_i$) of each node, as presented in \hyperref[eq:aggregation-func]{Equation~\ref*{eq:aggregation-func}}. As it is theoretically based, the attention layer tries to assign different importance to each edge through the attention scores. The encodings from a neighbour are scaled by the attention scores and then aggregated together. This scaling practically leads to different contributions of neighbour nodes. After the aggregation, applying the $\sigma$ as an activation function ensures the non-linearity.

\begin{equation}
	\label{eq:aggregation-func}
	\overrightarrow{h}^\prime_i = \sigma \bigg( \sum_{j \in \mathcal{N}_i} \alpha_{ij} \textbf{W}\overrightarrow{h}_j \bigg)
\end{equation}

Finally, to profit \textit{multi-head attention} mechanism, $L$ independent solitary attention mechanisms perform the presented transformation in \hyperref[eq:aggregation-func]{Equation~\ref*{eq:aggregation-func}}. According to \hyperref[eq:multi-head-attention]{Equation~\ref*{eq:multi-head-attention}}, averaging the results and finally applying a softmax yields the final prediction, which is produced by the multi-head attention network;

\begin{equation}
	\label{eq:multi-head-attention}
	\overrightarrow{h}^\prime_i =  \sigma \bigg( {1 \over L} \sum_{l=1}^{L} \sum_{j \in \mathcal{N}_i} \alpha_{ij}^l \textbf{W}^l\overrightarrow{h}_j \bigg)
\end{equation}
where $\alpha_{ij}^l$ refers to the attention scores produced by $l$-th attention mechanism, namely $a^l$; and $\textbf{W}^l$ refers to the corresponding weight matrix after the linear transformation. This was what happened in each layer.

\paragraph{Enriching the Knowledge Graph Attention Model by Knowledge Graph Embedding}
\label{sec:KGEmbedding}~\\
KGrAt-Net allows the users to profit from the embeddings of the knowledge graphs and the knowledge graph attention networks to carry out classification. It provides this supplementary option to enrich the classification and acquire another auxiliary representation of the pruned input graph. The graph is essentially made up of RDFs. In this step, the pruned knowledge graph will be transformed into a vector space, and an \textit{embedding matrix} will be yielded, in which its rows denote the graph's nodes and its columns, which denote the acquired dimensions after the embedding. In this investigation, the knowledge graph was embedded according to the procedure proposed by Ristoski et al. \cite{2019rdf2vec}. Their study proposed RDF2vec for embedding the RDF graphs and achieving the equivalent and more meaningful vector representation. In practice, RDF2vec efforts to maximally conserve the graph's structure, albeit that it fulfils dimensionality reduction on it. RDF2vec is basically inspired by the word2vec \cite{2013WORD2VEC}, a popular word embedding method that transforms words into a numerical vector space. It almost operates like word2vec. The sole difference is in their input sequence. Specifically, word2vec receives a set of sentences as the input sequence for training the model, whilst RDF2vec performs random walks on the graph to create sequences of RDF nodes to feed them into the learning model as the input sequence. As a result of such embedding, similar nodes locate close to each other in the final vector space and dissimilar ones do not. Exactly like what happens during word embedding in word2vec. Further details of RDF2vec are described in \hyperref[subsec: APP]{section~\ref*{subsec: APP}}

At the end of the third phase, a quick overview of this phase would be helpful. As it can also be seen in \hyperref[Fig: General architecture]{Figure~\ref*{Fig: General architecture}} one can conclude that:

\begin{enumerate}[label=(\roman*)]
	\setlength{\itemsep}{0pt}
	\setlength{\parskip}{0pt}
	\item input: pruned equivalent knowledge graph of the input text;
	\item output: predicted label of input text;
	\item objective: developing an attention network over the knowledge graph and enriching it by knowledge graph embedding to perform text classification.
\end{enumerate}

\subsection{Automatic Personality Prediction Using KGrAt-Net}
\label{sec:APPusingKGrAtNet}
At the moment, a knowledge graph attention network text classifier, namely KGrAt-Net, is available. To answer the research questions, we performed automatic personality prediction using KGrAt-Net, as described below.

\subsubsection{Dataset}
In the current investigation, we used Essays Dataset \cite{pennebaker1999} for training and testing KGrAt-Net. It comprises 2,467 essays written by psychology students. Afterwards, they were asked to fill out the Big Five Inventory Questionnaire. According to their responses, for each essay a binary label was assigned to each of five personality traits (OCEAN).

\begin{figure}
	\centering
	\includegraphics[width=0.7\textwidth]{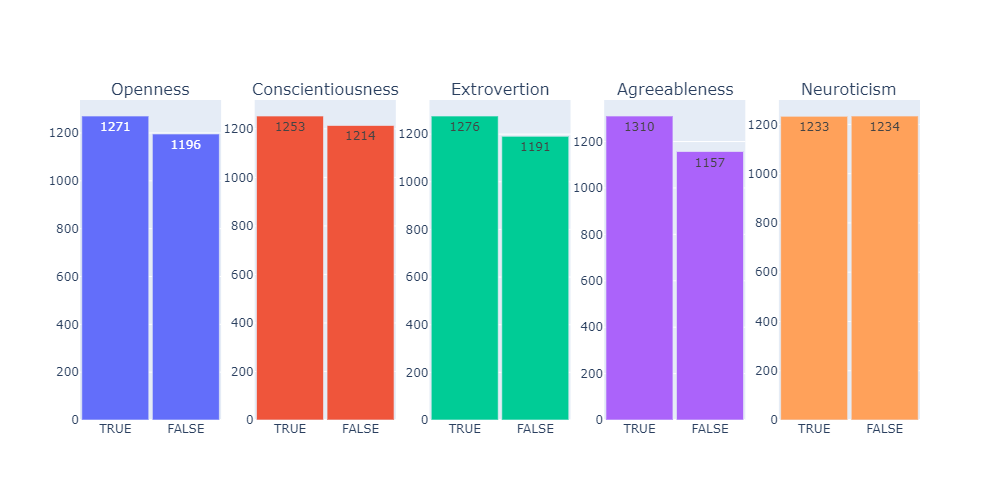} 
	\caption{The distribution of labels in each five personality traits in Essays Dataset}
	\label{Fig:Dataset label Distributions}
\end{figure}

\hyperref[Fig:Dataset label Distributions]{Figure~\ref*{Fig:Dataset label Distributions}} provides clear statistics for the distribution of labels in each personality trait in the dataset. The slight difference between the number of True and False labelled essays in each personality trait acknowledges that the dataset is balanced. Moreover, \hyperref[table:EssaysCorelation]{Table~\ref*{table:EssaysCorelation}} declares that there is no correlations between personality traits in Essays Dataset.

\begin{table}
	\centering
	\caption{The correlation matrix between five personality traits in Essays Dataset}
	\label{table:EssaysCorelation}
	\scriptsize
	
	\begin{adjustbox}{max width=\textwidth}
		
		\begin{tabular}{c c c c c c c c c c c}
			\toprule
			
			\textbf{Traits} &  & \textbf{O} &  & \textbf{C} &  & \textbf{E} &  & \textbf{A} &  & \textbf{N} \\
			\cmidrule{1-1}\cmidrule{3-3}\cmidrule{5-5}\cmidrule{7-7}\cmidrule{9-9}\cmidrule{11-11}
			O &   & 1 &   & - 0.027 &   & 0.079 &   & 0.018 &   & - 0.047  \\
			
			C &   &  &   & 1 &   & 0.14 &   & 0.13 &   & - 0.15  \\
			
			E &   &  &   &  &   & 1 &   & 0.12 &   & - 0.16  \\
			
			A &   &  &   &  &   &  &   & 1 &   & - 0.19 \\ 
			
			N &   &  &   &  &   &  &   &    &  &  1 \\
			
			\bottomrule
		\end{tabular}
		
	\end{adjustbox}
	
\end{table}

\subsubsection{Personality Prediction}
\label{subsec: APP}
To train KGrAt-Net, one should iteratively perform the first and second phases for each sample in the dataset. Consequently, a concise knowledgeable equivalent (namely the corresponding pruned knowledge graph) will be available for each sample in the dataset. We did it for each essay in Essays Dataset and acquired the equivalent pruned knowledge graphs. In practice, KGrAt-Net will be trained and tested using the equivalent pruned knowledge graph of each sample in the Essays Dataset. 

As described in \hyperref[sec:Knowledge Graph Preparation]{section~\ref*{sec:Knowledge Graph Preparation}}, KGrAt-Net performs the predictions as an \textit{essay\_nodes} classification problem. Each node was equipped with a feature vector (\overrightarrow{h}) and the target labels. More specifically, each \textit{essay\_node} contains: 

\begin{enumerate}[label=\roman*)]
	\setlength{\itemsep}{0pt}
	\setlength{\parskip}{0pt}
	\item a binary feature vector (\overrightarrow{h}) possessing the \textit{$<$entity\_attributes$>$} that demonstrates that each entity in the ``entity vocabulary" (the collection of all the entities that appeared in the Essays Dataset) is present (indicated by 1), or absent (indicated by 0) in the corresponding essay;
	\item  the \textit{$<$OCEAN\_labels$>$} a five-digit binary string that demonstrates personality traits' target labels for each essay (1 and 0 declare each essay's belonging and not belonging to each of the \{O, C, E, A, and N\} traits).
\end{enumerate}

\begin{figure}
	\centering
	\includegraphics[width=\textwidth]{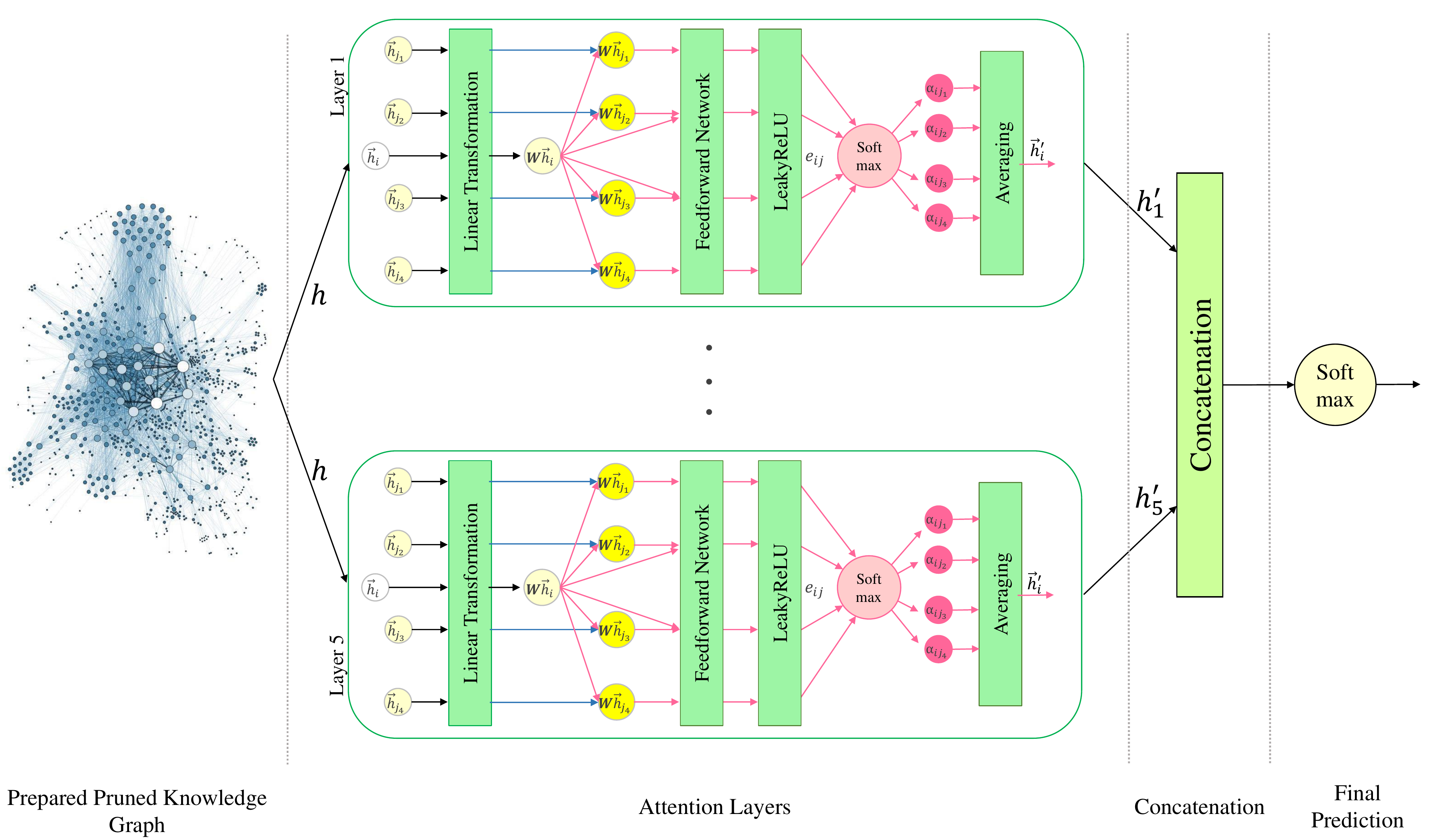} 
	\caption{The architecture of knowledge graph attention network classifier. The pruned knowledge graph is fed into the attention layers as a set of node features $h = \{\overrightarrow{h}_1, \overrightarrow{h}_2, ..., \overrightarrow{h}_N\}$. Each layer, computes the corresponding set of features after applying attention mechanism, namely $h^\prime = \{\overrightarrow{h}^\prime_1, \overrightarrow{h}^\prime_2, ..., \overrightarrow{h}^\prime_N\}$. Finally, the outputted set of node features from each layer (namely $h^\prime_l , l \in \{1, 2, ..,5\}$), are concatenated to produce the final prediction.}
	\label{Fig:Attention Network Architecture}
\end{figure}

During learning and testing the model, the set of node features $h = \{\overrightarrow{h}_1, \overrightarrow{h}_2, ..., \overrightarrow{h}_N\}, \overrightarrow{h}_i \in {\{0, 1\}}^F,$ were fed into the KGrAt-Net's attention model. Subsequently, KGrAt-Net build the model following the computations described in \hyperref[sec:Attention Network]{section~\ref*{sec:Attention Network}}. \hyperref[Fig:Attention Network Architecture]{Figure~\ref*{Fig:Attention Network Architecture}} presents the attention model which was used to perform the personality prediction.

To provide more flexibility, KGrAt-Net allows the user to customize some settings for learning the model, including the number of final classes, number of attention layers, the number of hidden units in each attention layer, number of units (neurons) in starting dense layers, number of attention heads in each multi-head attention layer, optimizer (as well as the learning rate), loss function, train-test split ratio, number of epochs, batch size, the application of early stopping (as well as the patience value), and the validation split ratio in each epoch.

It should also be mentioned that, since the difference between single-label and multi-label APP models is not significant \cite{Farandi2016}, the prediction in each five personality traits was fulfilled individually and independently of the others. Namely, the prediction was performed individually in each of the five \{O, C, E, A, and N\} personality traits. The correlation matrix between five personality traits (\hyperref[table:EssaysCorelation]{Table~\ref*{table:EssaysCorelation}}), also justifies this.

\begin{figure}
	\centering
	\includegraphics[width=0.8\textwidth]{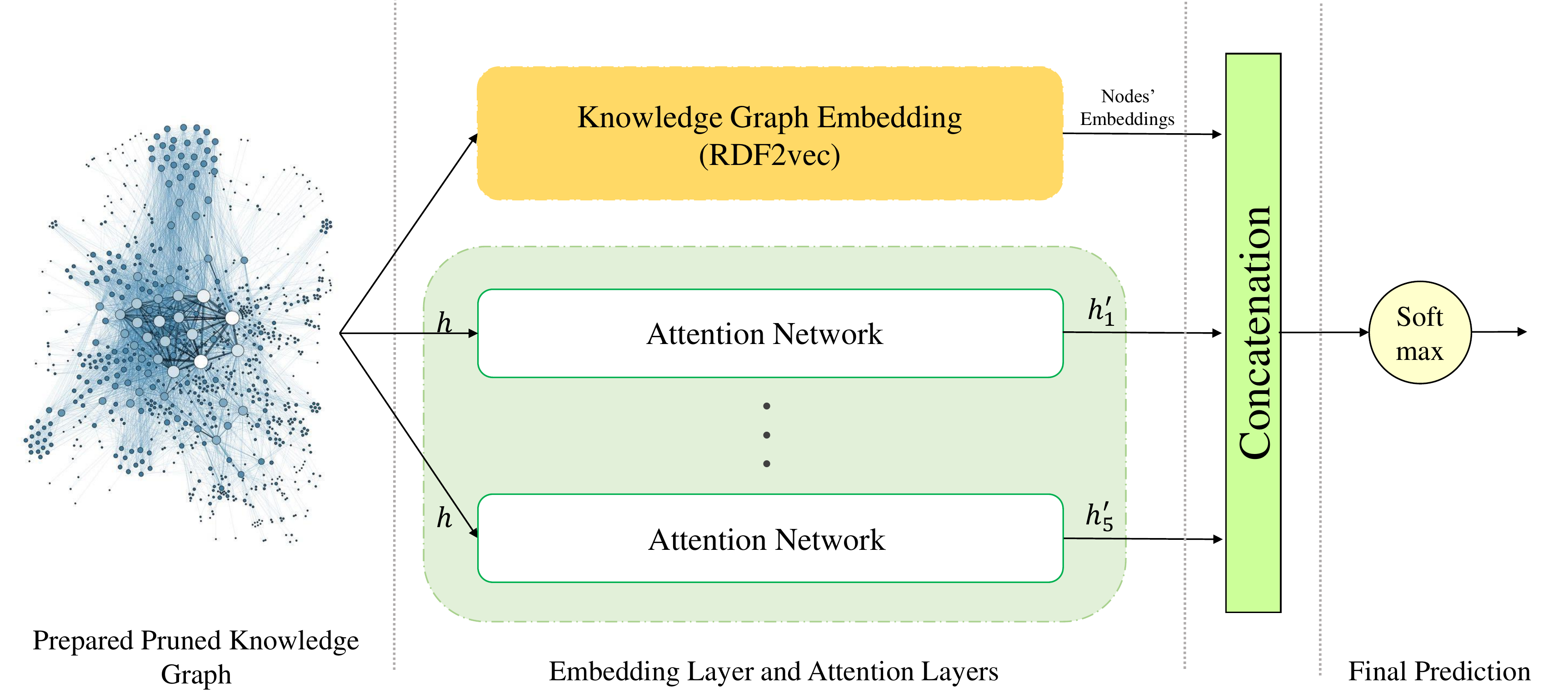} 
	\caption{The architecture of knowledge graph attention network classifier which is enriched by knowledge graph embedding. The predictions perform individually in each of the five \{O, C, E, A, and N\} personality traits.}
	\label{Fig:AttentionModelandKGEmbedding}
\end{figure}

Furthermore, to enrich the acquired representation, KGrAt-Net provides an option to benefit the pruned knowledge graph's embedding along with the attention mechanism (as described in \hyperref[sec:KGEmbedding]{section~\ref*{sec:KGEmbedding}}). Selecting this option, KGrAt-Net will be enriched by knowledge graph embedding. That is to say, the graph embedding will be concatenated to the attention networks results and then will be fed into the softmax classifier, as shown in \hyperref[Fig:AttentionModelandKGEmbedding]{Figure~\ref*{Fig:AttentionModelandKGEmbedding}}. KGrAt-Net, allows the user to configure the settings for RDF2vec. Specifying the maximum depth in each walk, the maximum number of walks per node to perform the random walks, and the embedding size is necessary. We set the maximum depth in each walk and the maximum number of walks per node, equal to 5, and the embedding size equal to 500.

Algorithm \ref{Algorithm_1_KGrAtNetPerformance} details a step-by-step flow of KGrAt-Net's classification method that would assist toward a better comprehension of its performance.

\begin{algorithm}
	\caption{The step-by-step flow of automatic personality prediction using KGrAt-Net}
	\label{Algorithm_1_KGrAtNetPerformance}
	\scriptsize
	
	\begin{algorithmic}[1]
		\MyFor{$e_{i} \in$ Essays Dataset}
		\STATEx \textcolor{gray}{$\; \; \; \%$ Description: Phase-1: Perform preprocessing activities for $e_{i}$, including:}
		\State \parbox[t]{360pt}{Perform tokenization}
		\State \parbox[t]{360pt}{Perform noise removal (punctuations, signs, and stop words)}
		\State \parbox[t]{360pt}{Perform normalization (lowercasing and lemmatization)}
		\State \parbox[t]{360pt}{Perform Named Entity Recognition (NER)}
		\State \parbox[t]{380pt}{Perform final preparations (duplicate element removal, first letter capitalization and whitespace replacement)}
		\STATEx \textcolor{gray}{$\; \; \; \%$ Description: Phase-2: Perform knowledge representation for $e_{i}$ , more specifically:} 
		\State \parbox[t]{360pt}{Build the corresponding knowledge graph for $e_{i}$ through DBpedia}
		\State \parbox[t]{360pt}{Perform knowledge graph pruning}
		\EndMyFor
		\STATEx
		\State Aggregate all of the pruned knowledge graphs and build an aggregated knowledge graph
		\STATEx
		\STATEx \textcolor{gray}{\% Description: knowledge graph preparation}
		\MyFor{$e_{i} \in$ Essays Dataset}
		\State Append corresponding \textit{essay\_node} to the knowledge graph
		\State \parbox[t]{360pt}{Connect the \textit{essay\_node} to \textit{entity\_nodes} which were present in the knowledge graph (in case of the occurrence of each \textit{entity\_node} in current $e_{i}$, add an edge between the corresponding \textit{essay\_node} and \textit{entity\_node})}
		\EndMyFor
		\STATEx
		\STATEx \textcolor{gray}{\% Description: automatic personality prediction}
		\MyFor{personality trait $t \in \{O, C, E, A, N\}$}
		\STATEx \textcolor{gray}{$\; \; \; \; \%$ Description: if prediction will be performed using attention mechanism + knowledge graph embedding}
		\IF{\textit{(Applying knowledge graph embedding while classification} $==$  \textit{True})}
		\STATE \parbox[t]{360pt}{Embed the knowledge graph and acquire corresponding embedding matrix}
		\MyFor{\textit{essay\_node} in knowledge graph}
		\STATE \parbox[t]{360pt}{Apply the attention mechanism over the knowledge graph and then acquire the corresponding feature vectors ($\overrightarrow{h}^\prime_i$) outputted from each attention layer}
		\STATE \parbox[t]{360pt}{Concatenate all of the feature vectors ($\overrightarrow{h}^\prime_i$) as well as the corresponding embedding vector}
		\STATE \parbox[t]{360pt}{Feed the concatenated vector to a softmax classifier to predict the personality trait ($t$)}
		\EndMyFor
		\STATEx \textcolor{gray}{$\; \; \; \; \%$ Description: if prediction will be performed just using the attention mechanism}
		\ELSE
		\MyFor{\textit{essay\_node} in knowledge graph}
		\STATE \parbox[t]{360pt}{Apply the attention mechanism over the knowledge graph and then acquire the corresponding feature vectors ($\overrightarrow{h}^\prime_i$) outputted from each attention layer}
		\STATE \parbox[t]{360pt}{Concatenate all of the feature vectors ($\overrightarrow{h}^\prime_i$)}
		\STATE \parbox[t]{360pt}{Feed the concatenated vector to a softmax classifier to predict the personality trait ($t$)}
		\EndMyFor
		\ENDIF
		\EndMyFor
	\end{algorithmic}
\end{algorithm}

\section{Results}
\label{sec:result}

\subsection{Evaluation Metrics}
Conventionally, classification models are evaluated using some eminent evaluation metrics, namely \textit{precision}, \textit{recall}, \textit{f-measure}, and \textit{accuracy}. Their values are determined by matching the elements of two sets: the set of essays' ``actual labels'' (which is sometimes referred to as \textit{gold standard}) and the set of essays' ``system predicted labels''. Matching each essay's actual binary label with its corresponding predicted label will lead us to one of the following situations:

\begin{enumerate}[label=\roman*)]
	\setlength{\itemsep}{0pt}
	\setlength{\parskip}{0pt}
	\item the actual label is true and the system predicted label is also true (which is referred to as \textit{TP} that indicates True Positive);
	\item the actual label is false and the system predicted label is also false (which is referred to as \textit{TN} that indicates True Negative);	
	\item the actual label is false while the system predicted label is true (which is referred to as \textit{FP} that indicates False Positive);	
	\item the actual label is true while the system predicted label is false (which is referred to as \textit{FN} that indicates False Negative).
\end{enumerate}

When evaluating an APP system, it is worth knowing what proportion of all predictions is predicted correctly. The system's correct predictions just occur when the essay's system predicted label is equal to its actual label. TP and TN specify this situation. Therefore, the system's correct predictions are equal to TP+TN. Besides, the total number of system predictions is equal to TP+TN+FP+FN. In consequence the proportion of systems correct predictions, which is known as \textit{accuracy}, is equal to $(TP + TN)/ (TP + TN + FP + FN)$.

What is more, some facets of the performance of a classification system reveal with precision, recall and their weighted harmonic mean, namely f-measure. If someone wants to know that, what proportion of system's true labeled predictions (TP+FP), has actual true labels (TP), \textit{precision} (\textit{P}) exactly discloses it; in fact, $P = TP/(TP + FP)$. Furthermore, If someone wants to know that, what proportion of the actual true labels (TP + FN), are predicted by the system (TP), \textit{recall} (\textit{R}) exactly discloses it; in fact $R = TP/(TP + FN)$. 

It should be noted that, despite their useful information, the precision and recall metrics can not be relied on individually to evaluate the performance of a classification system. Since there may be some cases with high values of precision and low values of recall simultaneously or contrariwise. It is basically due to their incomplete coverage of reports. Accordingly, \textit{f-measure} $= (2\times P\times R)/(P + R)$ is proposed to combine their included reports and solve the problem. However, as can be seen from the precision and recall, since the f-measure ignores all of the correctly false labelled samples by the system (namely, TN), it loses the ground to accuracy while evaluating an APP system. Hence, accuracy is preferred to f-measure for system evaluation.

\subsection{Evaluation Results}
We initiated this research to call into question the application of the attention mechanism over the corresponding knowledge graph for a given text in APP. To do so, we proposed KGrAt-Net, a three-phase text classifier. As discussed in the \hyperref[sec:KGRATNET]{fourth section}, receiving a text, in the first phase KGrAt-Net attempts to clean up the input text and transform it into a more digestible form for the machine by using some preprocessing activities. Next, in the second phase, it tries to represent the input text throughout building the corresponding knowledge graph. In this phase, it also attempts to prune unessential parts of the knowledge graph. Then finally, in the third phase, KGrAt-Net provides two ways to classify the input texts; one that includes a knowledge graph attention network and another that includes a knowledge graph attention network along with the knowledge graph's embedding. Our findings are described below. The parameter settings which were applied in our proposed method, are presented in \hyperref[table:ParameterSetting]{Table~\ref*{table:ParameterSetting}}.

\begin{table}
	\centering
	\caption{The parameter settings of the proposed APP method}
	\label{table:ParameterSetting}
	\scriptsize
	
	\begin{adjustbox}{max width=\textwidth}
		
		\begin{tabular}{r c}
			\toprule
			\textbf{Parameter} & \textbf{Setting}
			\\
			\midrule
			Train-Test split ratio (\%) & 80-20
			\\
			Number of epochs & 50
			\\
			Number of attention layers & 5
			\\
			Number of attention heads & 8
			\\
			Number of hidden units & 128
			\\
			Optimizer & Adam
			\\
			Learning rate & 3e-4
			\\
			Loss function & Binary crossentropy
			\\
			Early stopping & Applied on validation accuracy
			\\
			Patience value & 10
			\\
			Batch size & 32
			\\
			Cross validation & 10-fold
			\\
			\bottomrule
		\end{tabular}
		
	\end{adjustbox}
	
\end{table}

Some information about the final knowledge graph (which is acquired after the aggregation of all individual pruned knowledge graphs and knowledge graph preparation) may be useful. It contains 36,212 nodes (including 33,745 \textit{entity\_nodes} and 2,467 \textit{essay\_nodes}). It also possesses 632,941 edges (including 133,250 \textit{entity\_entity\_edges} and  499,691 \textit{essay\_entity\_edges}). Since there are some entity\_nodes that have appeared in each essay (refer to knowledge graph building and pruning in \ref{sec:KGBuilding} and \ref{sec:KGPruning}), there are always some edges between each \textit{essay\_nodes} and some \textit{entity\_nodes} in the final knowledge graph.

\hyperref[table:Evaluation_results_attention]{Table~\ref*{table:Evaluation_results_attention}} presents the results obtained from the evaluation of KGrAt-Net, when APP was just performed using knowledge graph attention network. It encompasses the values of four evaluation metrics, namely precision, recall, f-measure and accuracy for all of the OCEAN traits. As can be seen from the  \hyperref[table:Evaluation_results_attention]{Table~\ref*{table:Evaluation_results_attention}}, an average value of 64.82\% for precision indicates that, on average 64.82\% of the KGrAt-Net's true labelled predictions, were predicted correctly, while using knowledge graph attention network for APP. Furthermore, an average value of 81.44\% for recall denotes that KGrAt-Net in average has recalled (predicted) 81.44\% of the true labelled essays in the dataset correctly, in all of the five traits. On top of that, on average 70.26\% of KGrAt-Net's predictions in all of the five traits were accurate.

In a same manner, \hyperref[table:Evaluation_results_attentionAndKGEmbedding]{Table~\ref*{table:Evaluation_results_attentionAndKGEmbedding}} presents the results obtained from the evaluation of KGrAt-Net, when APP was performed using knowledge graph attention networks along with the knowledge graphs' embeddings. As can be seen from the  \hyperref[table:Evaluation_results_attentionAndKGEmbedding]{Table~\ref*{table:Evaluation_results_attentionAndKGEmbedding}}, KGrAt-Net in average of 69.27\%, predicted the personality traits of the input texts in all the five traits, precisely. In other words, as precision denotes, on average 69.27\% of the KGrAt-Net's true labelled predictions were predicted correctly. In addition, KGrAt-Net in average has recalled (predicted) 80.58\% of the true labelled essays in the dataset correctly in all of the five traits. Besides, on average 72.41\% of KGrAt-Net's predictions in all of the five traits were accurate.

\begin{table}
	\centering
	\caption{Evaluation results for APP using KGrAt-Net when classification is performed using knowledge graph attention network}
	\label{table:Evaluation_results_attention}
	\scriptsize
	
	\begin{adjustbox}{max width=\textwidth}
		
		\begin{tabular}{c c c c c c c c c c c c c}
			\toprule
			
			\textbf{Metrics} (\%) &  & \textbf{O} &  & \textbf{C} &  & \textbf{E} &  & \textbf{A} &  & \textbf{N} &  & \textbf{Avg.}\\
			\cmidrule{1-1}\cmidrule{3-3}\cmidrule{5-5}\cmidrule{7-7}\cmidrule{9-9}\cmidrule{11-11}\cmidrule{13-13}
			Precision &   & 67.93 &   & 67.22 &   & 64.94 &   & 61.43 &   & 62.60 &  & 64.82 \\
			
			Recall &   & 80.74 &   & 81.05 &   & 88.38 &   & 81.08 &   & 75.93 &  & 81.44 \\
			
			f-measure &   & 73.78 &   & 73.49 &   & 74.87 &   & 69.90 &   & 68.62 &  & 72.13 \\
			
			\midrule
			
			Accuracy &   & 71.60 &   & 70.59 &   & 70.99 &   & 68.56 &   & 69.57 &  & 70.26 \\ 
			\bottomrule
		\end{tabular}
		
	\end{adjustbox}
	
\end{table}

\begin{table}
	\centering
	\caption{Evaluation results for APP using KGrAt-Net when classification is performed using knowledge graph attention network along with knowledge graph embedding}
	\label{table:Evaluation_results_attentionAndKGEmbedding}
	\scriptsize
	
	\begin{adjustbox}{max width=\textwidth}
		
		\begin{tabular}{c c c c c c c c c c c c c}
			\toprule
			
			\textbf{Metrics} (\%) &  & \textbf{O} &  & \textbf{C} &  & \textbf{E} &  & \textbf{A} &  & \textbf{N} &  & \textbf{Avg.}\\
			\cmidrule{1-1}\cmidrule{3-3}\cmidrule{5-5}\cmidrule{7-7}\cmidrule{9-9}\cmidrule{11-11}\cmidrule{13-13}
			Precision &   & 70.45 &   & 72.95 &   & 73.86 &   & 66.90 &   & 62.17 &  & 69.27 \\
			
			Recall &   & 80.08 &   & 80.38 &   & 82.78 &   & 79.83 &   & 79.81 &  & 80.58 \\
			
			f-measure &   & 74.96 &   & 76.48 &   & 78.07 &   & 72.80 &   & 69.89 &  & 74.44 \\
			
			\midrule
			
			Accuracy &   & 72.21 &   & 73.43 &   & 74.24 &   & 71.20 &   & 70.99 &  & 72.41 \\ 
			\bottomrule
		\end{tabular}
		
	\end{adjustbox}
	
\end{table}

\subsection{Baseline Models}
To validate the performance of KGrAt-Net effectively, we compare it with the following state-of-the-art baselines, which were performed on Essays Dataset: 

\begin{itemize}
	
	\item \textbf{Tighe et al.}\cite{tighe2016perso}: they have used LIWC features to perform APP on Essays Dataset using several classifiers (like SVM, Sequential Minimal Optimization, and Linear Logistic Regression). They concentrated on removing insignificant LIWC features during classification by applying Information Gain and Principal Components Analysis (PCA).
	
	\item \textbf{Majumder et al.} \cite{Majumder2017}: they have proposed a CNN in which the document-level Mairesse features were fed into the model. They have trained five independent identical binary classifiers for OCEAN personality traits. 
	
	\item \textbf{Yuan et al.} \cite{YuanCu2018}: they have exploited deep features extracted through a CNN and then combined them with LIWC features to improve APP.
		
	\item \textbf{El-Demerdash et al.} \cite{ELDEMERDASH2020}: they have proposed the application of Universal Language Model Fine-Tuning (ULMFiT) for APP, which is an effective transfer learning method that can be applied in different language processing tasks. 
	
	\item \textbf{Jiang et al.} \cite{Jiang2020}: they have exploited pre-trained contextual embeddings (BERT and RoBERTa) to achieve more accurate predictions in APP.
	
	\item \textbf{Kazameini et al.} \cite{kazameini2020}: they have also suggested an APP system was based on BERT to extract contextualized word embeddings from textual data. Then, the embeddings and several psycholinguistic features, were fed into a Bagged-SVM classifier.
	
	\item \textbf{Wang et al.} \cite{Wang2020Encoding}: they have proposed a graph convolutional neural network that models users as well as their textual information (word-level and document-level) by a graph. Finally, the embeddings of this information were classified using a softmax classifier. 
	
	\item \textbf{Ramezani et al.} \cite{ramezani2021}: they have ensembled five inherently different classification models to profit their abilities in prediction simultaneously, including term frequency vector-based, ontology-based, enriched ontology-based, LSA-based, and BiLSTM-based methods.
	
	\item \textbf{Xue et al.} \cite{xue2021semantic}: they have proposed a semantic-enhanced APP system which acquires hierarchical semantic representations of the text elements.
	
	\item \textbf{El-Demerdash et al.} \cite{ELDASH2021}: they have proposed a deep learning-based APP system that was based on data-level and classifier-level fusion, which exploits various levels of information to improve the performance of an APP system.
	
	\item \textbf{Ramezani et al.} \cite{ramezani2022KGEnabeled}: they have proposed a knowledge graph-enabled APP system that classifies the input text by building, enriching, and embedding its corresponding knowledge graph.

\end{itemize}

The \hyperref[table:Baselines_Comparison]{Table~\ref*{table:Baselines_Comparison}}, provides an insight into the performance of baseline models in APP as well as the performance of proposed classification strategies by KGrAt-Net. As can be seen from the data in this table, KGrAt-Net's enriched classifier has achieved the best results for accuracy and f-measure in all five personality traits. Meanwhile, the unenriched classifier has also outperformed all baseline models in accuracy and f-measure except Ramezani et al. \cite{ramezani2022KGEnabeled}, in which it has just outperformed in O and N traits.

\begin{sidewaystable}
	\caption{Comparing the performance of baseline models in APP from text, which were performed on Essays Dataset (missing values are not reported).}
	
	\renewcommand{\arraystretch}{1.7}    
	\begin{tabular}{c c c c c c c c c c c c c c}
		\toprule
			
		\label{table:Baselines_Comparison}
		
		&\multicolumn{6}{c}{\textbf{f-measure} (\%)} &  & \multicolumn{6}{c}{\textbf{Accuracy} (\%)} \\\cmidrule{2-7} \cmidrule{9-14}
		Baselines & \textbf{O} & \textbf{C} & \textbf{E} &  \textbf{A} & \textbf{N} & \textbf{Avg.} &  & \textbf{O} & \textbf{C} & \textbf{E} &  \textbf{A} & \textbf{N} & \textbf{Avg.}\tabularnewline
		\midrule
		Tighe et al. \cite{tighe2016perso} & 61.90 & 56.00 & 55.60 & 55.70 & 58.30 & 57.50 &  & 61.95 & 56.04 & 55.75 & 57.54 &  58.31 & 57.92 \tabularnewline
		
		Majumder et al. \cite{Majumder2017} &   &   &   &  &  &  &  & 62.68 & 57.30 & 58.09 & 56.71 & 59.38 & 58.83 \tabularnewline
		
		Yuan et al. \cite{YuanCu2018}  &  &  &  &  &  &  &  & 62.00 & 57.00 & 58.00 & 56.00 & 59.00  & 58.40 \tabularnewline
		
		El-Demerdash et al. \cite{ELDEMERDASH2020}  &  &  &  &  &  &  &  & 63.30 & 57.97 & 58.85 & 59.25 & 59.88  & 59.85 \\
		
		Jiang et al. \cite{Jiang2020}  &  &  &  &  &  &  &  & 65.86 & 58.55 & 60.62 & 59.72 & 61.04  & 61.16 \\
		
		Kazameini et al. \cite{kazameini2020}   &  &  &  &  &  &  &  & 62.09 & 57.84 & 59.30 & 56.52 &  59.39  & 59.03 \\
		
		Wang et al. \cite{Wang2020Encoding}  & 67.00 & 68.00 & 67.00 & 69.00 & 69.00 & 68.00 &  & 64.80 & 59.10 & 60.00 & 57.70 & 63.00  & 60.92 \\
		
		Ramezani et al. \cite{ramezani2021}  & 57.37 & 59.74 & 65.80 & 61.62 & 60.69 & 61.04 &  & 56.30 & 59.18 & 64.25 & 60.31 &  61.14 & 60.24 \\
		
		Xue et al. \cite{xue2021semantic}  & 67.84  & 63.46  & 71.50  & 71.92 & 62.36 & 67.42 &  & 63.16 & 57.49 & 58.91 & 57.49 &  59.51  & 59.31  \\
		
		El-Demerdash et al. \cite{ELDASH2021}  &  &  &  &  &  &  &  & 65.60 & 59.52 & 61.15 & 60.80 &  62.20 & 61.85 \\
		
		Ramezani et al. \cite{ramezani2022KGEnabeled} & 73.64 & 75.68 & 77.72 & 71.78 & 68.34 & 73.43 &  & 71.40 & 72.62 & 73.83 & 70.18 &  69.37 & 71.48 \\
		
		
		KGrAt-Net \scriptsize(\textit{KG attention}) & 73.78 & 73.49 & 74.87 & 69.90  & 68.62 & 72.13 &  & 71.60 & 70.59 & 70.99 & 68.56 &  69.57 & 70.26 \\
		
		KGrAt-Net \scriptsize(\textit{KG attention+KG embedding}) & \textbf{74.96} & \textbf{76.48} & \textbf{78.08} & \textbf{72.80}  & \textbf{69.89} & \textbf{74.44} &  & \textbf{72.21} & \textbf{73.43} & \textbf{74.24} & \textbf{71.20} &  \textbf{70.99} & \textbf{72.41} \\
		
		\bottomrule
	\end{tabular}
	
\end{sidewaystable}

\section{Discussions}
\label{sec:Discussion}
This study was designed to assess the effect of the knowledge graph attention network in APP. To this end, we proposed KGrAt-Net, a knowledge graph attention network classifier that relies on applying an attention mechanism over the equivalent knowledge graph of text documents. Accordingly, it provides two classification strategies: the first performs APP by using a knowledge graph attention network classifier, and the second performs it by enriching the first strategy. More specifically, the second classification strategy uses a knowledge graph attention network along with knowledge graph embedding to perform classification.

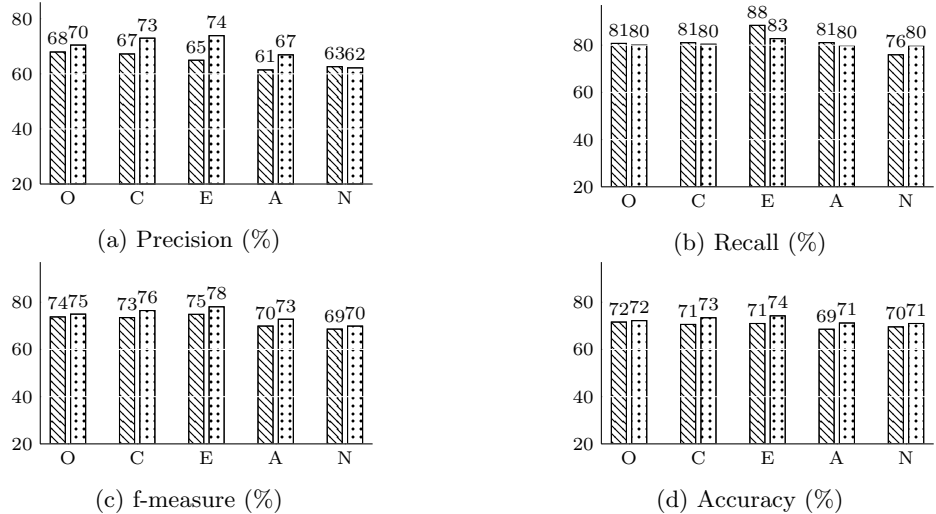
\begin{figure}
	\scriptsize
	
	\begin{subfigure}{.5\textwidth}
		\centering
		\begin{tikzpicture}
			\centering
			\begin{axis}[
				ybar, axis on top,
				height=4cm, width=6cm,
				bar width=0.2cm,
				ymajorgrids, tick align=inside,
				major grid style={draw=white},
				enlarge y limits={value=.1,upper},
				ymin=20, ymax=80,
				axis x line*=bottom,
				axis y line*=left,
				y axis line style={opacity=100},
				tickwidth=0pt,
				enlarge x limits=true,
				legend style={
					draw=none,   
					at={(0.5,-0.1)},
					legend cell align=left,
					anchor=north,
					legend columns=4,	
					/tikz/every even column/.append style={column sep=0.5cm}
				},
				symbolic x coords={
					O,C,E,A,N},
				xtick=data,
				nodes near coords={
					\pgfmathprintnumber[precision=0]{\pgfplotspointmeta}
				}
				]
				\addplot [black,postaction={pattern=north west lines}] coordinates {	
					(O, 67.93)
					(C, 67.22) 
					(E, 64.94)
					(A, 61.43) 
					(N, 62.60) };
				\label{pgf:AttenShape}
				\addplot [black, postaction={pattern= dots}] coordinates {	
					(O, 70.45)
					(C, 72.95) 
					(E, 73.86)
					(A, 66.90) 
					(N, 62.17)};
				\label{pgf:AttenEmbedShape}				
			\end{axis}
		\end{tikzpicture}
		\caption{Precision (\%)}
		\label{SUBFIGURE:Precision}
	\end{subfigure}
	\begin{subfigure}{.5\textwidth}
		\centering
		\begin{tikzpicture}
			\centering
			\begin{axis}[
				ybar, axis on top,
				height=4cm, width=6cm,
				bar width=0.2cm,
				ymajorgrids, tick align=inside,
				major grid style={draw=white},
				enlarge y limits={value=.1,upper},
				ymin=20, ymax=90,
				axis x line*=bottom,
				axis y line*=left,
				y axis line style={opacity=100},
				tickwidth=0pt,
				enlarge x limits=true,
				legend style={
					draw=none,   
					at={(0.5,-0.1)},
					legend cell align=left,
					anchor=north,
					legend columns=4,	
					/tikz/every even column/.append style={column sep=0.5cm}
				},
				symbolic x coords={
					O,C,E,A,N},
				xtick=data,
				nodes near coords={
					\pgfmathprintnumber[precision=0]{\pgfplotspointmeta}
				}
				]
				\addplot [black,postaction={pattern=north west lines}] coordinates {	
					(O, 80.74)
					(C, 81.05) 
					(E, 88.38)
					(A, 81.08) 
					(N, 75.93) };
				\addplot [black, postaction={pattern= dots}] coordinates {	
					(O, 80.08)
					(C, 80.38) 
					(E, 82.78)
					(A, 79.83) 
					(N, 79.81)};
				
			\end{axis}
		\end{tikzpicture}
		\caption{Recall (\%)}
		\label{SUBFIGURE:Recall}
	\end{subfigure}
	\begin{subfigure}{.5\textwidth}
		\centering
		\begin{tikzpicture}
			\begin{axis}[
				ybar, axis on top,
				height=4cm, width=6cm,
				bar width=0.2cm,
				ymajorgrids, tick align=inside,
				major grid style={draw=white},
				enlarge y limits={value=.1,upper},
				ymin=20, ymax=90,
				axis x line*=bottom,
				axis y line*=left,
				y axis line style={opacity=100},
				tickwidth=0pt,
				enlarge x limits=true,
				legend style={
					draw=none,   
					at={(0.5,-0.1)},
					legend cell align=left,
					anchor=north,
					legend columns=4,	
					/tikz/every even column/.append style={column sep=0.5cm}
				},
				symbolic x coords={
					O,C,E,A,N},
				xtick=data,
				nodes near coords={
					\pgfmathprintnumber[precision=0]{\pgfplotspointmeta}
				}
				]
				
				\addplot [black,postaction={pattern=north west lines}] coordinates {	
					(O, 73.78)
					(C, 73.49) 
					(E, 74.87)
					(A, 69.90) 
					(N, 68.62) };
				\addplot [black, postaction={pattern= dots}] coordinates {	
					(O, 74.96)
					(C, 76.48) 
					(E, 78.07)
					(A, 72.80) 
					(N, 69.89)};
			\end{axis}
		\end{tikzpicture}
		\caption{f-measure (\%)}
		\label{SUBFIGURE:fmeasure}
	\end{subfigure}
	\begin{subfigure}{.5\textwidth}
		\centering
		\begin{tikzpicture}
			\begin{axis}[
				ybar, axis on top,
				height=4cm, width=6cm,
				bar width=0.2cm,
				ymajorgrids, tick align=inside,
				major grid style={draw=white},
				enlarge y limits={value=.1,upper},
				ymin=20, ymax=90,
				axis x line*=bottom,
				axis y line*=left,
				y axis line style={opacity=100},
				tickwidth=0pt,
				enlarge x limits=true,
				legend style={
					draw=none,   
					at={(0.5,-0.1)},
					legend cell align=left,
					anchor=north,
					legend columns=4,	
					/tikz/every even column/.append style={column sep=0.5cm}
				},
				symbolic x coords={
					O,C,E,A,N},
				xtick=data,
				nodes near coords={
					\pgfmathprintnumber[precision=0]{\pgfplotspointmeta}
				}
				]
				\addplot [black,postaction={pattern=north west lines}] coordinates {	
					(O, 71.60)
					(C, 70.59) 
					(E, 70.99)
					(A, 68.56) 
					(N, 69.57) };
				\addplot [black, postaction={pattern= dots}] coordinates {	
					(O, 72.21)
					(C, 73.43) 
					(E, 74.24)
					(A, 71.20) 
					(N, 70.99)};
			\end{axis}
		\end{tikzpicture}
		\caption{Accuracy (\%)}
		\label{SUBFIGURE:Accuracy}
	\end{subfigure}
	\caption{The comparison of four evaluation metrics for both classification techniques provided by KGrAt-Net, namely knowledge graph attention network classification (depicted by \ref{pgf:AttenShape}) and knowledge graph attention network along with knowledge graph embedding classification (depicted by \ref{pgf:AttenEmbedShape}). The values are rounded.}
	\label{fig:KGratNETMethodsComparison}
\end{figure}

A comparison of four evaluation metrics, for both of the classification strategies proposed by KGrAt-Net, is shown in \hyperref[fig:KGratNETMethodsComparison]{Figure~\ref*{fig:KGratNETMethodsComparison}}. From the \hyperref[SUBFIGURE:Accuracy]{accuracy chart (d)} in this figure, and from the data in \hyperref[table:Evaluation_results_attention]{Table~\ref*{table:Evaluation_results_attention}} and \hyperref[table:Evaluation_results_attentionAndKGEmbedding]{Table~\ref*{table:Evaluation_results_attentionAndKGEmbedding}} it is apparent that the second strategy, performs APP more accurate than the first strategy, in all of the Big Five personality traits. Enriching the knowledge graph attention network classifier with knowledge graph embedding concludes more accurate predictions than an unenriched classifier. That is to say, combining the nodes' feature vectors with nodes' embedding vectors leads to better results rather than when the nodes' feature vectors are used solely. In the same manner, the second classification strategy has obtained better results than the first one, in all of the five traits for  \hyperref[SUBFIGURE:fmeasure]{f-measure (c)}. 

From another perspective, as discussed earlier, this study aimed to assess the importance of substituting text elements with more knowledgeable and knowing-full alternatives for the machine during APP computations. Considering these results, it was clearly confirmed that more knowledgeable and more knowing-full representations of the input text conclude more accurate results in APP. More specifically, paying different attentions to the neighbour concepts (nodes) in the knowledge graph during decision making in KGrAt-Net's first classification strategy increases the effect of more prominent concepts during classification and concludes more accurate decisions. This idea was also confirmed by KGrAt-Net's second classification strategy when the knowledge graph attention classifier was enriched by knowledge graph embedding. That is to say, enriching the attention mechanism's feature vectors by knowledge graphs embedding vectors would cause even more accurate decisions. Indeed, since the embedding methods also try to increase the effect of more prominent concepts through dimensionality reduction during decision makings, they also attempt to utilize a more knowledgeable and more knowing-full representation of input text. To paraphrase, more knowings generally will bring more accurate predictions. What is more, each of these representations is capable of acquiring the knowings in certain aspects, and combining (enriching) them would cover more knowings during decision making and would diminish the effect of neglected information by each of the methods during decision making.

\begin{sidewaysfigure}
	\caption{Comparing the \textit{accuracy} of baseline models in each OCEAN traits, with KGrAt-Net's two proposed classifiers.}
	\label{fig:AccuracyComparisonInBaselineModels}
	\centering
	\scriptsize
	\begin{subfigure}[b]{0.48\textwidth}
		\centering
		\begin{tikzpicture}
			\begin{axis}[
				xbar,
				xmin=55,
				enlargelimits=0.15,
				height=0.55\textwidth,
				bar width=3pt,
				ytick={1,2,3,4,5,6,7,8,9,10,11,12,13},
				yticklabels={KGrAt-Net (KG att.+KG emb.),KGrAt-Net (KG att.),Ramezani et al. \cite{ramezani2022KGEnabeled}, El-Demerdash et al. \cite{ELDASH2021}, Xue et al. \cite{xue2021semantic}, Ramezani et al. \cite{ramezani2021}, Wang et al. \cite{Wang2020Encoding}, Kazameini et al. \cite{kazameini2020}, Jiang et al. \cite{Jiang2020}, El-Demerdash et al. \cite{ELDEMERDASH2020}, Yuan et al. \cite{YuanCu2018}, Majumder et al. \cite{Majumder2017}, Tighe et al. \cite{tighe2016perso}}, 
				yticklabel style={text width=3.4cm, align=right, font=\tiny},
				width={.8\textwidth}, 
				nodes near coords, 
				nodes near coords align={horizontal},
				]
				\addplot coordinates {(72.21,1) (71.60,2) (71.40,3) (65.60,4) (63.16,5) (56.30,6) (64.80,7) (62.09,8) (65.86,9) (63.30,10) (62.00,11) (62.68,12) (61.95,13)};
			\end{axis}
		\end{tikzpicture}
		\caption{Openness (O)}
		\label{fig:BaselinesCompareO}
	\end{subfigure}	
	\hfill
	\begin{subfigure}[b]{0.48\textwidth}
		\centering
		\begin{tikzpicture}
			\begin{axis}[
				xbar,
				xmin=55,
				enlargelimits=0.15,
				height=0.55\textwidth,
				bar width=3pt,
				ytick={1,2,3,4,5,6,7,8,9,10,11,12,13},
				yticklabels={KGrAt-Net (KG att.+KG emb.),KGrAt-Net (KG att.),Ramezani et al. \cite{ramezani2022KGEnabeled}, El-Demerdash et al. \cite{ELDASH2021}, Xue et al. \cite{xue2021semantic}, Ramezani et al. \cite{ramezani2021}, Wang et al. \cite{Wang2020Encoding}, Kazameini et al. \cite{kazameini2020}, Jiang et al. \cite{Jiang2020}, El-Demerdash et al. \cite{ELDEMERDASH2020}, Yuan et al. \cite{YuanCu2018}, Majumder et al. \cite{Majumder2017}, Tighe et al. \cite{tighe2016perso}}, 
				yticklabel style={text width=3.4cm, align=right, font=\tiny},
				width={.8\textwidth}, 
				nodes near coords, 
				nodes near coords align={horizontal},
				]
				\addplot coordinates {(73.43,1) (70.59,2) (72.62,3) (59.52,4) (57.49,5) (59.18,6) (59.10,7) (57.84,8)  (58.55,9) (57.97,10) (57.00,11) (57.30,12) (56.04,13)};
			\end{axis}
		\end{tikzpicture}
		\caption{Conscientiousness (C)}
		\label{fig:BaselinesCompareC}
	\end{subfigure}
	\vskip\baselineskip	
	\begin{subfigure}[b]{0.48\textwidth}
		\centering
		\begin{tikzpicture}
			\begin{axis}[
				xbar,
				xmin=55,
				enlargelimits=0.15,
				height=0.55\textwidth,
				bar width=3pt,
				ytick={1,2,3,4,5,6,7,8,9,10,11,12,13},
				yticklabels={KGrAt-Net (KG att.+KG emb.),KGrAt-Net (KG att.),Ramezani et al. \cite{ramezani2022KGEnabeled}, El-Demerdash et al. \cite{ELDASH2021}, Xue et al. \cite{xue2021semantic}, Ramezani et al. \cite{ramezani2021}, Wang et al. \cite{Wang2020Encoding}, Kazameini et al. \cite{kazameini2020}, Jiang et al. \cite{Jiang2020}, El-Demerdash et al. \cite{ELDEMERDASH2020}, Yuan et al. \cite{YuanCu2018}, Majumder et al. \cite{Majumder2017}, Tighe et al. \cite{tighe2016perso}}, 
				yticklabel style={text width=3.4cm, align=right, font=\tiny},
				width={.8\textwidth}, 
				nodes near coords, 
				nodes near coords align={horizontal},
				]
				\addplot coordinates {(72.21,1) (71.60,2) (71.40,3) (65.60,4) (63.16,5) (56.30,6) (64.80,7) (62.09,8) (65.86,9) (63.30,10) (62.00,11) (62.68,12) (61.95,13)};
			\end{axis}
		\end{tikzpicture}
		\caption{Extroversion (E)}
		\label{fig:BaselinesCompareE}
	\end{subfigure}
	\hfill	
	\begin{subfigure}[b]{0.48\textwidth}
		\centering
		\begin{tikzpicture}
			\begin{axis}[
				xbar,
				xmin=55,
				enlargelimits=0.15,
				height=0.55\textwidth,
				bar width=3pt,
				ytick={1,2,3,4,5,6,7,8,9,10,11,12,13},
				yticklabels={KGrAt-Net (KG att.+KG emb.),KGrAt-Net (KG att.),Ramezani et al. \cite{ramezani2022KGEnabeled}, El-Demerdash et al. \cite{ELDASH2021}, Xue et al. \cite{xue2021semantic}, Ramezani et al. \cite{ramezani2021}, Wang et al. \cite{Wang2020Encoding}, Kazameini et al. \cite{kazameini2020}, Jiang et al. \cite{Jiang2020}, El-Demerdash et al. \cite{ELDEMERDASH2020}, Yuan et al. \cite{YuanCu2018}, Majumder et al. \cite{Majumder2017}, Tighe et al. \cite{tighe2016perso}}, 
				yticklabel style={text width=3.4cm, align=right, font=\tiny},
				width={.8\textwidth}, 
				nodes near coords, 
				nodes near coords align={horizontal},
				]
				\addplot coordinates {(71.20,1) (68.56,2) (70.18,3) (60.80,4) (57.49,5) (60.31,6) (57.70,7) (56.52,8) (59.72,9) (59.25,10) (56.00,11) (56.71,12) (57.54,13)};
			\end{axis}
		\end{tikzpicture}
		\caption{Agreeableness (A)}
		\label{fig:BaselinesCompareA}
	\end{subfigure}	
	\vskip\baselineskip
	\begin{subfigure}[b]{0.48\textwidth}
		\centering
		\begin{tikzpicture}
			\begin{axis}[
				xbar,
				xmin=55,
				enlargelimits=0.15,
				height=0.55\textwidth,
				bar width=3pt,
				ytick={1,2,3,4,5,6,7,8,9,10,11,12,13},
				yticklabels={KGrAt-Net (KG att.+KG emb.),KGrAt-Net (KG att.),Ramezani et al. \cite{ramezani2022KGEnabeled}, El-Demerdash et al. \cite{ELDASH2021}, Xue et al. \cite{xue2021semantic}, Ramezani et al. \cite{ramezani2021}, Wang et al. \cite{Wang2020Encoding}, Kazameini et al. \cite{kazameini2020}, Jiang et al. \cite{Jiang2020}, El-Demerdash et al. \cite{ELDEMERDASH2020}, Yuan et al. \cite{YuanCu2018}, Majumder et al. \cite{Majumder2017}, Tighe et al. \cite{tighe2016perso}}, 
				yticklabel style={text width=3.4cm, align=right, font=\tiny},
				width={.8\textwidth}, 
				nodes near coords, 
				nodes near coords align={horizontal},
				]
				\addplot coordinates {(70.99,1) (69.57,2) (69.37,3) (62.20,4) (59.51,5) (61.14,6) (63.00,7) (59.39,8) (61.04,9) (59.88,10) (59.00,11) (59.38,12) (58.31,13)};
			\end{axis}
		\end{tikzpicture}
		\caption{Neuroticism (N)}
		\label{fig:BaselinesCompareN}
	\end{subfigure}
\end{sidewaysfigure}
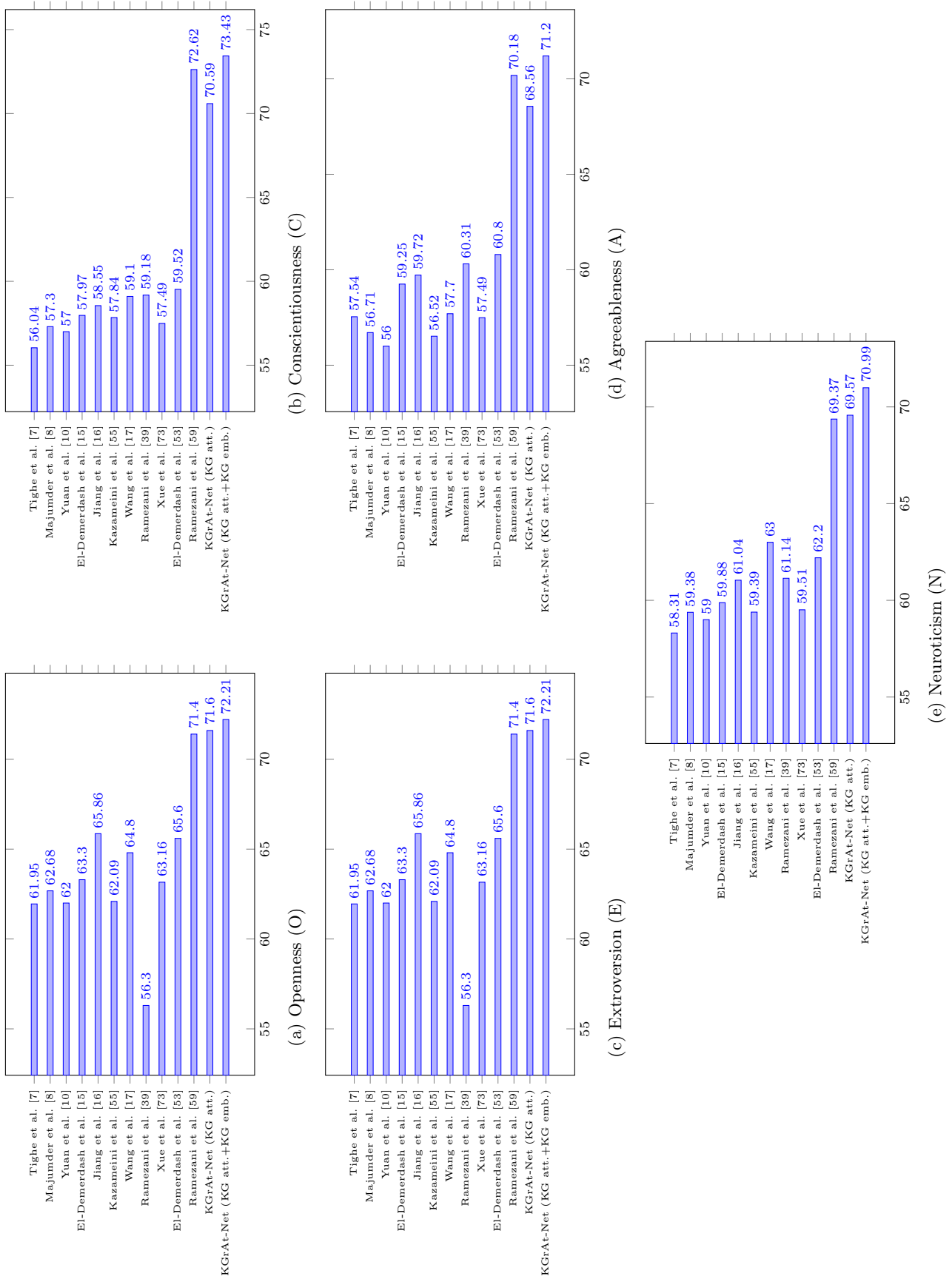

Comparing the accuracy of KGrAt-Net's proposed classification strategies with the baseline models (as can be seen in \hyperref[fig:AccuracyComparisonInBaselineModels]{Figure~\ref*{fig:AccuracyComparisonInBaselineModels}}), shows that KGrAt-Net has considerably enhanced the performance of APP. Simply put, classification using a knowledge graph attention network has significantly improved the accuracy of APP (as the average accuracies in \hyperref[table:Baselines_Comparison]{Table~\ref*{table:Baselines_Comparison}} approves that). More precisely, KGrAt-Net's first classification strategy has outperformed all of the state-of-the-art baseline models, except in some traits in Ramezani et al. \cite{ramezani2022KGEnabeled} (namely, C, E, and A). Their proposed APP system was basically based on knowledge graph embedding. An implication of this is the fact that, in spite of the knowledge graph attention network's successes in predictions, text-based APP using knowledge graph embedding causes more accurate predictions in some traits. This is clearly acknowledged by KGrAt-Net's enriched classification strategy when the knowledge graph attention network was enriched with knowledge graph embedding. The \hyperref[fig:AccuracyComparisonInBaselineModels]{Figure~\ref*{fig:AccuracyComparisonInBaselineModels}} clearly compares the accuracy of KGrAt-Net's proposed strategies with the state of the art baseline models in APP.

Eventually, we are going to answer the research questions (as expressed in \hyperref[Sec:Introduction]{Introduction}) according to our findings as follows:

\begin{itemize}
	\item[\bfseries RQ.1] The results of our investigations revealed that the knowledge graph attention network has a significant positive effect on text-based APP. Comparing the average accuracies between the first classification strategy proposed by KGrAt-Net and baseline models in \hyperref[table:Baselines_Comparison]{Table~\ref*{table:Baselines_Comparison}} clearly denotes the fact that there are considerable differences between the average accuracies in almost all baseline models (except Ramezani et al. \cite{ramezani2022KGEnabeled} in which it has outperformed to KGrAt-Net in C, E, and A). Comparing the accuracies in each of the OCEAN traits which are depicted in \hyperref[fig:AccuracyComparisonInBaselineModels]{Figure~\ref*{fig:AccuracyComparisonInBaselineModels}} also justifies the ability of knowledge graph attention network in text-based APP. In fact, it has empowered the model to pay attention to the most relevant parts of the knowledge graph and hence has aided in making better decisions. Specifically, regarding the nature of the attention mechanism, it should be stated that assigning different importance values for the neighbour concepts (nodes) in the knowledge graph during feature transformation causes better utilization of determining concepts during APP.
	
	\item[\bfseries RQ.2] According to our findings (as can be seen in \hyperref[table:Baselines_Comparison]{Table~\ref*{table:Baselines_Comparison}} and \hyperref[fig:AccuracyComparisonInBaselineModels]{Figure~\ref*{fig:AccuracyComparisonInBaselineModels}}), enriching the knowledge graph attention network by knowledge graph embedding, substantially enhances the text-based APP accuracies inasmuch as it concludes the best results among all baseline models. In fact, before the enrichment (as the results of KGrAt-Net's first strategy illustrate), KGrAt-Net  despite defeating all baseline models, failed to overcome one of the baseline models (Ramezani et al. \cite{ramezani2022KGEnabeled}) in some traits (C, E, and A), which it is basically established on the basis of knowledge graph embedding. This obviously indicates that embedding the knowledge graph leads to benefit some neglected aspects of knowings that were missed by the attention mechanism. As a matter of fact, both the attention mechanism and embedding method over the knowledge graph remove certain parts of the graph, respectively, through paying attention to the most prominent parts of the graph and through dimensionality reduction. Hence, it should be stated that enriching the knowledge graph attention network by knowledge graph embedding brings some helpful knowings that improve APP's accuracy.
	
	\item[\bfseries RQ.3] Regarding the obtained results, applying the knowledge graph attention network generally enhances the performance of the text-based APP system in all of the OCEAN personality traits, although the enhancements are not quite equal. Being more specific, the knowledge graph attention network in practice has yielded the best results among the state-of-the-art baseline models in O and N, while in C, E, and A, defeating all other models, it has just fallen behind Ramezani et al. \cite{ramezani2022KGEnabeled} (as can be seen in \hyperref[table:Baselines_Comparison]{Table~\ref*{table:Baselines_Comparison}} and \hyperref[fig:AccuracyComparisonInBaselineModels]{Figure~\ref*{fig:AccuracyComparisonInBaselineModels}}). Similarly, enriching it by the knowledge graph embedding totally has improved the performance of predictions in all of the OCEAN traits, inasmuch as KGrAt-Net definitely has achieved the best results among the state of the art baseline models. Generally, it can be concluded that applying knowledge graph attention as well as knowledge graph embedding in APP, capably improves the accuracy of APP, independent of each of the five personality traits. Regarding the essence of applied methods, it would be because acquiring the knowledge behind words through knowledge graphs brings more knowledgeable and knowing-full representations, which are eligible substitutions that lead to better predictions in all of the five traits.
	
	\item[\bfseries RQ.4] The knowledge graph was the foundation of KGrAt-Net's classifiers as the representation of input texts. Perusing the resulting values for applied evaluation metrics, we can claim that the knowledge graph as an alternative representation of the input text, adroitly encompasses the necessary information for the machine to perform text-based APP. This claim is clearly supported by both of the classification strategies proposed by KGrAt-Net. Actually, in practice representing the input text using its equivalent knowledge graph has provided the competent structure both for applying the attention mechanism and for embedding it to vector space. Furthermore, as can be seen from Table \ref{table:Baselines_Comparison} among all of the text-based methods, those that are relied on knowledge graphs (namely, \cite{ramezani2022KGEnabeled} and the two proposed methods in the current study), have achieved the most striking results in accuracy and f-measure. This is an evident reason for the capability of knowledge graphs for representing the input text and providing eligible substitutions during APP.
	
	\item[\bfseries RQ.5] According to our experience, basically when we represent the input text using knowledge graphs, we will face very large graphs, even for small pieces of text. Indeed, it is anticipated since the knowledge graph entails all of the existing information about each of the appeared concepts in the input text, as well as the relationship among them. The huge amount of information in the knowledge graph (both for nodes and edges), imposes a great deal of time and memory complexity on the system and even, most of the time, it would be impossible to proceed with the task. In fact, in practice not all of the existing information in the knowledge graph, take a positive role in the current task. Hence, it should be noted that while working with knowledge graphs, one should regard this concern and find an appropriate strategy to cope with it.
	
\end{itemize}

\section{Conclusion}
\label{sec:conclusion}
The present study was designed to determine the effect of the knowledge graph attention networks on text-based automatic personality prediction (APP). To this end, we proposed KGrAt-Net, which is a knowledge graph attention network classifier. It follows a three-phase approach to perform classification: it carries out some essential preprocessing activities in the first phase that makes the input text ready for main processes in the next phase; then, in the second phase, trying to achieve the knowledge behind the presented concepts in the input text, it represents the input text knowledgeably through building its equivalent knowledge graph; finally in the third phase, it designs a classification model through applying an attention network over the acquired knowledge graph.

This is the first time that a knowledge graph attention network has been used to perform a text-based APP. The findings from this study make several noteworthy contributions to the current literature. First, the knowledge graph attention network significantly improves the accuracy of a text-based APP system. Second, enriching the knowledge graph attention network classifier with knowledge graph embedding would considerably enhance the accuracy of predictions. Third, this study has gone some way towards enhancing our understanding of the efficacy of knowledge graphs in representing the knowledge behind the existing concepts in the input text and providing an eligible alternative for the input text that encompasses more comprehensive and digestible information for the machine, rather than the input string of characters. Finally, it also sheds some light on the importance of knowledge representation for the machine while performing human-like decision making.

\section*{Competing interests}
\addcontentsline{toc}{section}{Competing interests}
The authors declare no competing interests.

\section*{Author information}
\addcontentsline{toc}{section}{Author information}

\subsection*{Affiliations}
\textbf{Computerized Intelligence Systems Laboratory, Department of Computer Engineering, Faculty of Electrical and Computer Engineering, University of Tabriz, Tabriz, Iran}

Majid Ramezani \& Mohammad-Reza Feizi-Derakhshi\\
\textbf{Department of Computer Engineering, Faculty of Electrical and Computer Engineering, University of Tabriz, Tabriz, Iran}

Mohammad-Ali Balafar

\subsection*{Contributions}
M.R. and M.R.F.D contributed the following: designed and performed the experiments, formal analysis, writing, and investigations.
M.R. and M.A.B contributed the following: review and editing manuscript, data curation, supervision and the analytical methods verification.

\subsection*{Corresponding authors}
Mohammad-Reza Feizi-Derakhshi and Majid Ramezani

\section*{Data availability}
\addcontentsline{toc}{section}{Data availability}
The datasets generated during the current study are available from the corresponding author on reasonable request.

\addcontentsline{toc}{section}{\refname}
\bibliographystyle{naturemag}\scriptsize  
\bibliography{MyBib.bib}

\end{document}